\documentclass[10pt,twocolumn,letterpaper]{article}

\usepackage{cvpr}              
\usepackage{chen}
\usepackage{soul}
\newcommand{\cmark}{\ding{51}}%
\newcommand{\xmark}{\ding{55}}%
\usepackage[pagebackref,breaklinks,colorlinks]{hyperref}

\definecolor{mygray2}{gray}{.6}
\definecolor{mywarning}{RGB}{233,144,61}
\definecolor{myred}{RGB}{192,0,0}
\definecolor{myyellow}{RGB}{255,192,0}
\definecolor{myorange}{RGB}{243,225,214} 

\def\cvprPaperID{2606}
\def\confName{CVPR}
\def\confYear{2022}

\begin{document}

\title{Locality-Aware Inter-and Intra-Video Reconstruction for\\ Self-Supervised Correspondence Learning }

\author{\!\!\!\!\!\!\!Liulei Li$^{1,6}$\thanks{Work done during an internship at Baidu Research.}~, Tianfei Zhou$^{2}$, Wenguan Wang$^{3}$\thanks{Corresponding author: \textit{Wenguan Wang}.}~, Lu Yang$^{4}$, Jianwu Li$^{1}$~, Yi Yang$^{5}$\\
\small{\!\!\!\!\!$^1$ Beijing Institute of Technology}~~\small{$^2$ ETH Zurich}~~\small{$^3$ ReLER, AAII, University of Technology Sydney}\\\small{$^4$ Beijing University of Posts and Telecommunications}~~\small{$^5$ CCAI, Zhejiang University}~~\small{$^6$ Baidu Research}\\
\small\url{https://github.com/0liliulei/LIIR}
}

\maketitle

\begin{abstract}
Our$_{\!}$ target$_{\!}$ is$_{\!}$ to$_{\!}$ learn$_{\!}$ visual$_{\!}$ correspondence$_{\!}$ from$_{\!}$ unla- beled videos. We develop \textsc{Liir}, a \underline{l}ocality-aware \underline{i}nter-and \underline{i}ntra-video$_{\!}$ \underline{r}econstruction$_{\!}$ method$_{\!}$ that$_{\!}$ fills$_{\!}$ in$_{\!}$ three$_{\!}$~miss- ing$_{\!}$  pieces,$_{\!}$ \ie,$_{\!}$ instance$_{\!}$ discrimination,$_{\!}$ location$_{\!}$ awareness,$_{\!}$ and$_{\!}$ spatial$_{\!}$ compactness,$_{\!}$ of$_{\!}$
self-supervised$_{\!}$ correspondence  learning  puzzle.  First,  instead  of  most  existing  efforts~focu- sing$_{\!}$ on$_{\!}$ intra-video$_{\!}$ self-supervision$_{\!}$ only,$_{\!}$ we$_{\!}$ exploit cross- video affinities as extra negative samples within a unified, inter-and$_{\!}$ intra-video$_{\!}$ reconstruction$_{\!}$ scheme.$_{\!\!}$ This$_{\!}$ enables$_{\!}$~in- stance$_{\!}$ discriminative$_{\!}$ representation$_{\!}$ learning$_{\!}$ by$_{\!}$~contrasting  desired$_{\!}$ intra-video$_{\!}$ pixel$_{\!}$ association$_{\!}$ against$_{\!}$ negative$_{\!}$ inter-video$_{\!}$ correspondence.$_{\!}$ Second,$_{\!}$ we$_{\!}$ merge$_{\!}$ position$_{\!}$~informa- tion$_{\!}$ into$_{\!}$ correspondence$_{\!}$ matching,$_{\!}$ and$_{\!}$ design$_{\!}$ a$_{\!}$ position shifting$_{\!}$ strategy$_{\!}$ to$_{\!}$ remove$_{\!}$ the$_{\!}$ side-effect$_{\!}$ of$_{\!}$ position encod- ing$_{\!}$ during$_{\!}$ inter-video$_{\!}$ affinity$_{\!}$ computation,$_{\!}$ making$_{\!}$~our$_{\!}$ \textsc{Liir} location-sensitive.$_{\!}$ Third,$_{\!}$ to$_{\!}$ make$_{\!}$ full$_{\!}$ use$_{\!}$ of$_{\!}$ the$_{\!}$ spatial$_{\!}$~conti- nuity nature of video data, we impose a compactness-based constraint$_{\!}$ on$_{\!}$ correspondence$_{\!}$ matching,$_{\!}$ yielding$_{\!}$ more$_{\!}$ spar- se$_{\!}$ and$_{\!}$ reliable$_{\!}$ solutions.$_{\!}$
The$_{\!}$ learned$_{\!}$~representation surpasses self-supervised state-of-the-arts on label propagation
tasks including objects, semantic parts, and keypoints.
\end{abstract}

\vspace{-20pt}
\section{Introduction}
	\vspace{-3pt}
$_{\!}$As$_{\!}$ a$_{\!}$ fundamental$_{\!}$ problem$_{\!}$ in$_{\!}$ computer$_{\!}$ vision,$_{\!}$ correspondence matching facilitates many applications, such as$_{\!}$~scene$_{\!}$ understanding$_{\!\!}$~\cite{satkin20133dnn},$_{\!\!}$ object$_{\!\!}$ dynamics$_{\!}$ modeling$_{\!\!}$~\cite{hu2018videomatch},$_{\!\!}$ and$_{\!}$ 3D reconstruction$_{\!}$~\cite{geiger2011stereoscan}.$_{\!}$ However,$_{\!}$ supervising$_{\!}$ representation$_{\!}$~for visual$_{\!}$ correspondence$_{\!}$ is$_{\!}$ not$_{\!}$ trivial,$_{\!}$ as obtaining pixel-level manual annotations is costly, and sometimes even prohibi- tive$_{\!}$ (due$_{\!}$ to occlusions and free-form object deformations). Although synthetic data would serve an alternative in some low-level visual correspondence tasks (\eg, optical flow$_{\!}$ estimation \cite{baker2011database}), they limit the generalization to real scenes.

Using$_{\!}$ natural$_{\!}$ videos$_{\!}$ as$_{\!}$ a$_{\!}$ source$_{\!}$ of$_{\!}$ free$_{\!}$ supervision,$_{\!}$~\ie, \textit{self-supervised$_{\!}$ temporal$_{\!}$ correspondence$_{\!}$ learning},$_{\!}$ is$_{\!}$~consi- dered$_{\!}$ as$_{\!}$ appealing$_{\!}$~\cite{lai2019self}. This is because videos$_{\!}$ contain$_{\!}$~rich realistic appearance  and shape variations with almost infinite supply, and deliver valuable supervisory signals from the intrinsic coherence, \ie, correlations among frames.

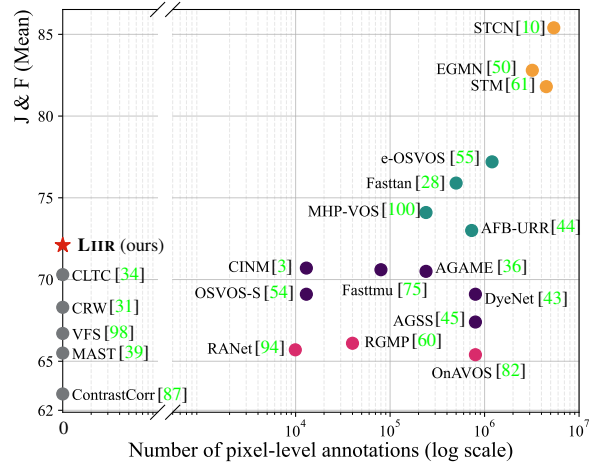
\begin{figure}[t]
	\vspace{-10pt}
	\begin{center}
		\includegraphics[width=0.95\linewidth]{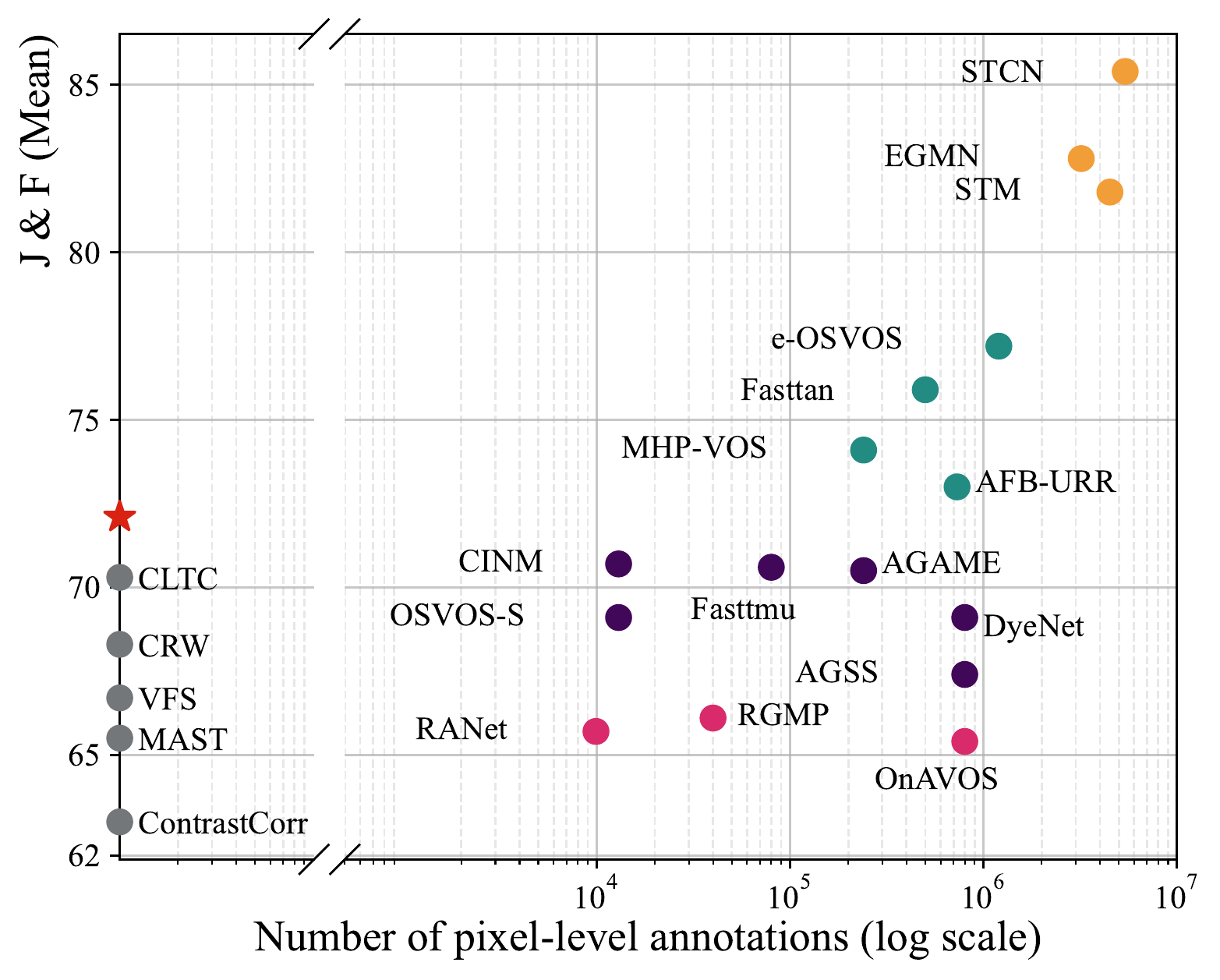}
		\put(-204.5,14){\scriptsize 0}
		\put(-197.6,83.2){\scriptsize \textbf{\textsc{Liir}} (ours)}
		\put(-168.08,26.82){\scriptsize \!~\cite{wang2021contrastive}}
		\put(-182.6,42.49){\scriptsize \!~\cite{lai2020mast}}	
		\put(-188.2,50.4){\scriptsize \!~\cite{xu2021rethinking}}	
		\put(-186.3,59.95){\scriptsize \!~\cite{jabri2020space}}
		\put(-184.2,72.24){\scriptsize \!~\cite{jeon2021mining}}
		\put(-131.2,44.62){\scriptsize \!~\cite{wang2019ranet}}
		\put(-127.80,65.45){\scriptsize \!~\cite{maninis2018video}}
		\put(-124.90,75.4){\scriptsize \!~\cite{bao2018cnn}}
		\put(-71.80,47.10){\scriptsize \!~\cite{oh2018fast}}
		\put(-63.00,55.5){\scriptsize \!~\cite{lin2019agss}}
		\put(-40.1,35.74){\scriptsize \!~\cite{voigtlaender2017online}}
		\put(-24.20,63.2){\scriptsize \!~\cite{li2018video}}
		\put(-39.6,75.05){\scriptsize \!~\cite{johnander2019generative}}
		\put(-77.3,66.7){\scriptsize \!~\cite{sun2020fast}}
		\put(-19.2,90.2){\scriptsize \!~\cite{liang2020video}}
		\put(-83.3,96.78){\scriptsize \!~\cite{xu2019mhp}}
		\put(-36.5,144.2){\scriptsize \!~\cite{oh2019video}}
		\put(-70.5,107.2){\scriptsize \!~\cite{huang2020fast}}
		\put(-57.4,116.6){\scriptsize \!~\cite{meinhardt2020make}}
		\put(-43.6,150.6){\scriptsize \!~\cite{lu2020video}}
		\put(-32.6,165.75){\scriptsize \!~\cite{cheng2021rethinking}}
		\end{center}
	\vspace{-22pt}
	\captionsetup{font=small}
	\caption{$_{\!}$\small\textbf{Performance$_{\!}$ comparison}$_{\!}$ over$_{\!}$ DAVIS$_{17\!}$ \texttt{val}.$_{\!}$ Our$_{\!}$ \textsc{Liir} surpasses all existing self-supervised methods, and is on par with many fully-supervised ones trained with massive annotations.
	}
	\label{fig:1}
	\vspace{-18pt}
\end{figure}

Along this direction, existing solutions are typically built upon a \textit{reconstruction} scheme (\ie,$_{\!}$ each$_{\!}$ pixel$_{\!}$ from$_{\!}$ a$_{\!}$ `query' frame is reconstructed by finding and assembling relevant pixels from adjacent frame(s)) \cite{vondrick2018tracking,lai2019self,lai2020mast}, and/or adopt a \textit{cycle-consistent tracking} paradigm (\ie, pixels/patches are encouraged to fall into the same location after one cycle of forward and backward tracking) \cite{wang2019unsupervised,wang2019learning,li2019joint,lu2020learning,jabri2020space}.

\begin{figure*}[t]
	\vspace{-2pt}
	\begin{center}
		\includegraphics[width=\linewidth]{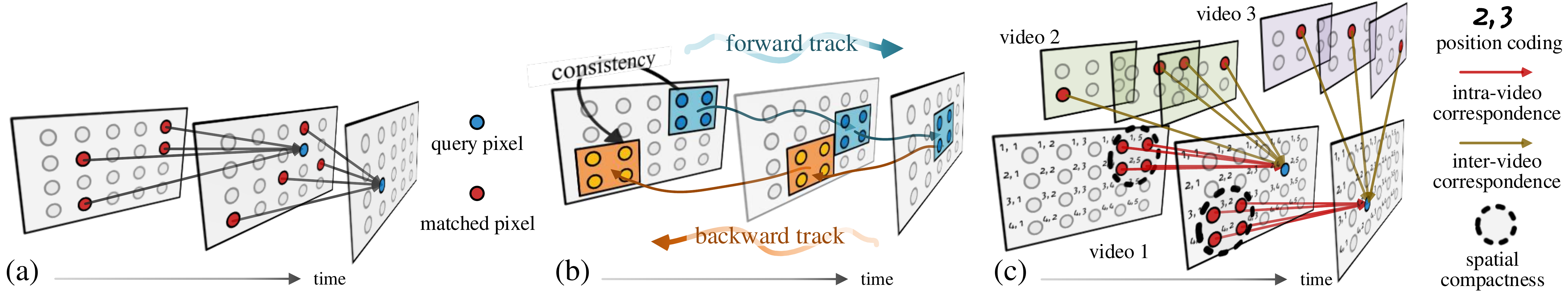}
		\end{center}
	\vspace{-18pt}
	\captionsetup{font=small}
	\caption{\small \textbf{Illustration of different self-supervised architectures for temporal correspondence learning} (\S\ref{sec:rw}): (a) reconstruction based, (b) cycle-consistency based, and (c) our \textsc{Liir} that addresses instance discrimination, location awareness, and spatial compactness.
	}
	\label{fig:2}
	\vspace{-13pt}
\end{figure*}

Unfortunately, these successful approaches largely neglect three crucial abilities for robust temporal correspondence learning, namely \textbf{instance discrimination}, \textbf{location awareness}, and \textbf{spatial compactness}. First, many of them share a narrow view that only considers intra-video context for correspondence learning. As it is hard to derive a free signal from a single video for identifying different object instances, the learned features are inevitably less instance-discriminative. Second, existing methods are typically built without explicit position representation. Such design seems counter-intuitive, given the extensive evidence that spatial position is encoded in human visual system~\cite{hess1992coding} and plays a vital role when human track objects~\cite{oksama2016position}.  Third, as the visual world is continuous and smoothly-varying, both spatial and temporal coherence naturally exist in videos. While numerous strategies are raised to address smoothness on the time axis, far less attention has been paid to the spatial case.

To fill in these three missing pieces to the puzzle~of~self-\\
\noindent supervised correspondence learning, we present a \underline{l}ocality- aware$_{\!}$ \underline{i}nter-and$_{\!}$ \underline{i}ntra-video$_{\!}$ \underline{r}econstruction$_{\!}$ framework$_{\!}$ --$_{\!}$ \textsc{Liir}.$_{\!\!}$ \textbf{First}, we augment existing intra-video analysis based correspondence learning strategy with \textit{inter-video context}, which is informative for instance-level separation. This leads~to~an  inter-and$_{\!}$ intra-video$_{\!}$ reconstruction$_{\!}$ based$_{\!}$ training$_{\!}$ objective, that \textit{inspires} intra-video positive correspondence matching,\\
\noindent  but$_{\!}$ \textit{penalizes}$_{\!}$ unreliable$_{\!}$ pixel$_{\!}$ associations$_{\!}$ within$_{\!}$ and$_{\!}$ cross videos.  We empirically verify that our inter-and intra-video\\
\noindent  reconstruction strategy can yield more discriminative features, that encode higher-level semantics beyond low-level intra-instance invariance modeled by previous algorithms. \textbf{Second}, to make our \textsc{Liir} more location-sensitive, we~learn to encode position information into the representation.~Al- though position bias is favored for intra-video correspon-   dence matching, it is undesired in the inter-video case.~We   thus devise a \textit{position} \textit{shifting} strategy to foster the~strength and$_{\!}$ circumvent$_{\!}$ the$_{\!}$ weaknesses$_{\!}$ of$_{\!}$ position$_{\!}$ encoding.$_{\!}$~We$_{\!}$~ex- perimentally$_{\!}$ show$_{\!}$ that,$_{\!}$ explicit$_{\!}$ position$_{\!}$ embedding$_{\!}$ benefits\\
\noindent  correspondence matching. \textbf{Third}, we involve a spatial compactness prior in intra-video pixel-wise affinity estimation, resulting in sparse yet compact associations.$_{\!}$ For$_{\!}$ each query pixel, the distribution of related pixels is fit by a mixture$_{\!}$ of$_{\!}$ Gaussians.$_{\!}$ This$_{\!}$ enforces$_{\!}$ each$_{\!}$ query$_{\!}$ pixel$_{\!}$ to$_{\!}$ match$_{\!}$~only$_{\!}$ a handful of spatially close pixels in an adjacent frame. Our experiments show that such compactness prior not only regularizes training, but also removes outliers during inference.

These three contributions together make \textsc{Liir} a powerful framework
for self-supervised correspondence learning. Without any adaptation, the learned representation is effec-\\
\noindent tive$_{\!}$ for$_{\!}$ various$_{\!}$ correspondence-related$_{\!}$ tasks,$_{\!}$ \ie,$_{\!}$ video$_{\!}$ object segmentation, semantic part propagation, pose tracking. On these tasks, \textsc{Liir} consistently outperforms unsupervised state-of-the-arts and is comparable to, or even better than, some task-specific fully-supervised methods (\eg, Fig.$_{\!}$~\ref{fig:1}).

	\vspace{-3pt}
\section{Related Work}\label{sec:rw}
	\vspace{-2pt}
\noindent\textbf{Self-Supervised Temporal Correspondence Learning.} In the$_{\!}$ video$_{\!}$ domain,$_{\!}$ correspondence$_{\!}$ matching$_{\!}$ plays$_{\!}$ a$_{\!}$ central role in many tasks (\eg, video segmentation$_{\!}$~\cite{hu2018videomatch}, flow esti- mation$_{\!}$~\cite{dosovitskiy2015flownet,ilg2017flownet}$_{\!}$ and$_{\!}$  object$_{\!}$ tracking$_{\!}$~\cite{bertinetto2016fully}).$_{\!}$ An$_{\!}$ emerging$_{\!}$ line of work tackles this problem in a self-supervised learning paradigm, by exploiting the temporal coherence in videos. One may group these work into two major classes. The$_{\!}$ first$_{\!}$ class$_{\!}$ of$_{\!}$ methods$_{\!}$~\cite{vondrick2018tracking,lai2019self,lai2020mast}$_{\!}$ poses$_{\!}$ a$_{\!}$ \textit{colorization} proxy task (Fig.$_{\!}$~\ref{fig:2}(a)), \ie, reconstruct a query frame from an adjacent$_{\!}$ frame,$_{\!}$ according$_{\!}$ to$_{\!}$ their$_{\!}$ correspondence.$_{\!}$ The$_{\!}$ latter type of methods$_{\!}$~\cite{wang2019unsupervised,wang2019learning,li2019joint,lu2020learning,jabri2020space}$_{\!}$ performs$_{\!}$ forward$_{\!}$ and$_{\!}$ backward$_{\!}$ tracking and penalizes the inconsistency between the start and end positions of the tracked$_{\!}$ pixels$_{\!}$ or$_{\!}$ regions$_{\!}$~(Fig.$_{\!}$~\ref{fig:2}(b)). The$_{\!}$ basic$_{\!}$ idea$_{\!}$ --$_{\!}$ \textit{cycle-consistency}$_{\!}$ --$_{\!}$ is$_{\!}$ also$_{\!}$ adopted$_{\!}$ in$_{\!}$~un- supervised$_{\!}$  tracking$_{\!}$~\cite{wang2019unsupervised,zheng2021learning},$_{\!}$ optical$_{\!}$ flow$_{\!}$ \cite{meister2018unflow,zou2018df}$_{\!}$ and$_{\!}$ depth$_{\!}$ estimation$_{\!~\!\!}$\cite{janai2018unsupervised,yin2018geonet}.$_{\!}$ Though impressive,$_{\!}$ these methods miss three key elements for robust correspondence matching: instance discrimination, location awareness, and spatial$_{\!}$ compactness. In response, \textsc{Liir} is equipped with three specific modules (Fig.$_{\!}$~\ref{fig:2}(c)). First, for instance discriminative~repre- sentation learning, it adopts an inter-and intra-video recons- truction$_{\!}$ scheme$_{\!}$ that$_{\!}$ formulates$_{\!}$ contrast$_{\!}$ over$_{\!}$ inter-and$_{\!}$ intra- video affinities. Second, it involves position encoding into representation$_{\!}$ learning.$_{\!}$ Third,$_{\!}$ it$_{\!}$ imposes$_{\!}$ a$_{\!}$ spatial$_{\!}$ compact- ness prior to both correspondence learning and inference.

$_{\!}$We$_{\!}$ are$_{\!}$ not$_{\!}$ the$_{\!}$ first$_{\!}$ to$_{\!}$ explore$_{\!}$ inter-video$_{\!}$ context.$_{\!}$~In$_{\!}$~\cite{lu2020learning},$_{\!}$ Lu \etal raise an unsupervised learning objective, \ie, discriminate between a set of surrogate video classes, but they compute this over video-level embeddings,$_{\!}$ not$_{\!}$~sufficient$_{\!}$ for pixel-wise correspondence$_{\!}$ learning.$_{\!}$ In$_{\!}$~\cite{wang2021contrastive}, Wang$_{\!}$~\etal$_{\!}$~also simultaneously consider inter-and intra-video representation associations, but requiring pre-aligned \textit{patch pairs}. In addition,$_{\!}$ they$_{\!}$ use$_{\!}$ three$_{\!}$ loss$_{\!}$ terms$_{\!}$ to$_{\!}$ address$_{\!}$ the$_{\!}$ desired$_{\!}$ intra-inter constraints,~which are, however, formulated as a single unified$_{\!}$ training$_{\!}$ objective$_{\!}$ in$_{\!}$ our$_{\!}$ case.$_{\!}$ Moreover,$_{\!}$ we$_{\!}$ take a further step by addressing location awareness and spatial compactness, instead inter-video context only. In$_{\!}$~\cite{xu2021rethinking}, Xu \etal revisit the idea of image-level similarity learning and consider frame pairs from the same videos are positive samples and pairs from different videos are negative, however, they find negative samples (\ie, inter-video context) hurt the\\
\noindent performance of their model. In contrast, we formulate inter-and intra-video context through a unified, pixel-wise affinity framework that boosts instance-level discrimination without sacrificing intra-instance invariance. Our results suggest that how to make a good use of negative samples for temporal correspondence learning is still an intriguing question.

\noindent\textbf{$_{\!}$Self-Supervised$_{\!}$ Video$_{\!}$ Representation$_{\!}$ Learning.}$_{\!}$ Corres- pondence$_{\!}$ learning$_{\!}$ approaches$_{\!}$ that$_{\!}$ use$_{\!}$ unlabeled video data fall$_{\!}$ in$_{\!}$ a$_{\!}$ broad$_{\!}$ field$_{\!}$ of$_{\!}$ self-supervised$_{\!}$ video representation learning. Towards learning transferable video representation, diverse pretext tasks are proposed to explore different intrinsic properties of videos as free supervisory signals, including temporal sequence ordering$_{\!}$~\cite{misra2016shuffle,fernando2017self,wei2018learning}, predicting motion patterns$_{\!}$~\cite{agrawal2015learning,tung2017self,pathak2017learning,gan2018geometry,wang2019self}, solving space-time cubic puzzles$_{\!}$ \cite{kim2019self}, anticipating future representations$_{\!}$~\cite{lotter2016deep,vondrick2016anticipating},  and temporally$_{\!}$ aligning$_{\!}$ videos$_{\!}$~\cite{dwibedi2019temporal}.$_{\!}$ The$_{\!}$ learned$_{\!}$ representations are$_{\!}$ compact$_{\!}$ video$_{\!}$ descriptors$_{\!}$ that$_{\!}$ can$_{\!}$ be$_{\!}$ generalized$_{\!}$~to$_{\!}$~va- rious$_{\!}$ downstream$_{\!}$ tasks$_{\!}$ (\eg,$_{\!}$ action$_{\!}$ recognition$_{\!}$ \cite{han2020memory,han2020self,wu2021contrastive,qian2021spatiotemporal},$_{\!}$ video$_{\!}$ captioning$_{\!}$~\cite{sun2019videobert,zhu2020actbert},$_{\!}$ video$_{\!}$ retrieval$_{\!}$~\cite{miech2020end}), while$_{\!}$ in$_{\!}$ this work we specifically focus on learning fine-grained visual features for pixel-level correspondence matching.

\noindent\textbf{Self-Supervised Image Representation Learning.} Basically, self-supervised approaches for image representation learning share a similar key idea with the counterparts for videos -- design pretext tasks so as to mine supervision signals from the inherent information inside images$_{\!}$~\cite{yang2021multiple}. Examples of such tasks include estimating spatial context$_{\!}$~\cite{doersch2015unsupervised}, predicting image rotations$_{\!}$~\cite{gidaris2018unsupervised}, solving jigsaw puzzles$_{\!}$~\cite{noroozi2016unsupervised}, among many others$_{\!}$~\cite{wang2015unsupervised,larsson2016learning,pathak2016context}. Recent efforts were mainly devoted to improving deep metric
learning techniques in large$_{\!}$ scale,$_{\!}$ \ie,$_{\!}$ adopt$_{\!}$ an$_{\!}$ instance$_{\!}$ discrimination$_{\!}$ pretext$_{\!}$~task where the cross-entropy objective is used to discriminate each image from other different images (\ie, negative sam- ples)$_{\!}$ \cite{oord2018representation,hjelm2019learning,he2020momentum,chen2020improved,chen2020simple}. In this work, we assimilate the idea of contrasting positive samples (intra-video affinities) over numerous negative ones (inter-video affinities) for robust temporal correspondence matching, and formulate this within a unified  inter-and intra-video reconstruction framework.

\noindent\textbf{Video Mask Propagation.} This task, a.k.a. semi-automatic video segmentation, aims at propagating first-frame object masks over the whole video sequence$_{\!}$~\cite{tsai2010,wang2018semi}. It addresses classic$_{\!}$ automatic$_{\!}$ video$_{\!}$ segmentation$_{\!}$ techniques'$_{\!}$~\cite{DBLP:conf/iccv/PapazoglouF13,DBLP:conf/cvpr/WangSP15,wang2019learning,lu2019see,zhou2021target,zhou2020motion}$_{\!}$ lack$_{\!}$ of$_{\!}$ flexibility$_{\!}$ in$_{\!}$ defining$_{\!}$ target$_{\!}$ objects$_{\!}$~\cite{wang2021survey}.$_{\!}$ Depending on how they make use of the first-frame supervision, existing mask propagation models can be classified into three groups: i) \textit{online fine-tuning} based methods that fine-tune a generic segmentation network with masks during inference$_{\!}$~\cite{caelles2017one,voigtlaender2017online}; and ii) \textit{matching} based methods that directly condition the segmentation network on the first-frame masks and/or previous segments$_{\!}$~\cite{DBLP:conf/cvpr/PerazziKBSS17,DBLP:conf/cvpr/JampaniGG17,DBLP:conf/iccv/YoonRKLSK17,oh2019video,ventura2019rvos,lu2020video}.
As shown in~Fig.$_{\!}$~\ref{fig:1}, extensive annotations are typically required to train such systems, \ie, first pre-train on ImageNet$_{\!}$~\cite{ILSVRC15} and then fine-tune with COCO$_{\!}$~\cite{lin2014microsoft}, DAVIS$_{\!}$~\cite{perazzi2016benchmark}, Youtube-VOS$_{\!}$~\cite{xu2018youtube}, \etc. In contrast, we pursue a more annotation-efficient solution; similar to previous self-supervised correspondence learning methods$_{\!}$~\cite{vondrick2018tracking,lai2019self,lu2020learning,lai2020mast}, \textsc{Liir} is trained using unlabeled videos only. Once trained, it can be directly applied for mask propagation, without adaptation.

	\vspace{-4pt}
\section{Our Approach}\label{sec:3}
	\vspace{-2pt}
We present \textsc{Liir}, a self-supervised framework that learns dense$_{\!}$ correspondence$_{\!}$ from$_{\!}$ raw$_{\!}$ videos.$_{\!}$ Before$_{\!}$ elaborating on$_{\!}$ our$_{\!}$ model$_{\!}$ design$_{\!}$ (\textit{cf}.~\S\ref{sec:3.2}),$_{\!}$ we$_{\!}$ first$_{\!}$ review$_{\!}$ the$_{\!}$ classic$_{\!}$~reconstruction based$_{\!}$ temporal$_{\!}$ correspondence$_{\!}$ learning$_{\!}$~stra- tegy (\textit{cf}.~\S\ref{sec:3.1}), which serves as the basis of our \textsc{Liir}.
	\vspace{-1pt}
\subsection{Preliminary: Learning Temporal Correspondence through Frame Reconstruction}\label{sec:3.1}
	\vspace{-1pt}
Due to the appearance continuity in video, one can~con- sider pixels in a `query' frame as being \textit{copied} from some locations$_{\!}$ of$_{\!}$ other$_{\!}$ `reference'$_{\!}$ frames.$_{\!}$ In light of this, a few studies$_{\!}$~\cite{vondrick2018tracking,lai2019self} raise$_{\!}$ a$_{\!}$ reconstruction-based$_{\!}$ correspondence learning scheme: each query pixel struggles to find pixels in a reference frame that can best reconstruct itself.

$_{\!}$Formally,$_{\!}$ let$_{\!}$ $I_{q}, I_{r\!}\!\in_{\!}\!\mathbb{R}^{H_{\!}\times_{\!}W_{\!}\times_{\!}3\!}$ respectively$_{\!}$ denote$_{\!}$ a$_{\!}$ query$_{\!}$ frame$_{\!}$ and$_{\!}$ a$_{\!}$ reference$_{\!}$ frame$_{\!}$ from$_{\!}$ the$_{\!}$ same$_{\!}$ video.$_{\!}$ They$_{\!}$ are projected into a pixel embedding space by a ConvNet encoder (\eg, ResNet\!~\cite{he2016deep}) $\phi_{\!}\!:_{\!}\! \mathbb{R}^{H_{\!}\times_{\!}W_{\!}\times_{\!}3\!}\!\rightarrow_{\!}\!\mathbb{R}^{h_{\!}\times_{\!}w_{\!}\times_{\!}c}$,$_{\!}$ such$_{\!}$ that$_{\!}$ $\bm{I}_{q}, \bm{I}_{r\!}\!=_{\!}\!\phi(I_{q}), \phi(I_{r})$.$_{\!}$ The$_{\!}$ \textit{copy} operator can be approximated as an inter-frame affinity matrix $A\!\in\![0,1]^{hw_{\!}\times_{\!}hw}$:
\vspace{-1pt}
\begin{equation}\small
    \begin{aligned}\label{eq:1}
A(i,j)\!=\!\frac{\exp(\bm{I}_{q}(i)\!\cdot\!\bm{I}_{r}(j))}{\textstyle\sum\nolimits_{j'}\exp(\bm{I}_{q}(i)\!\cdot\!\bm{I}_{r}(j'))}, ~i,j\!\in\!\{1, \cdots, hw\}\!\!\!
    \end{aligned}
    \vspace{-1pt}
\end{equation}
where $A(i,j)\!\in_{\!}\![0,1]$ refers to $(i,j)$-th element in $A$, signifying the similarity between pixel $i$ in ${I}_{q}$ and pixel $j$ in ${I}_{r}$, and `$\cdot$' stands for the dot product. In this way,  $A$ gives the strength of all$_{\!}$ the$_{\!}$ pixel$_{\!}$ pair-wise$_{\!}$ correspondence$_{\!}$ between$_{\!}$ $\bm{I}_{q\!}$ and$_{\!}$ $\bm{I}_{r}$,$_{\!}$ according$_{\!}$~to$_{\!}$ which pixel $i$ in ${I}_{q}$ can be reconstructed by a weighted sum of pixels in ${I}_{r}$:
\vspace{-1pt}
\begin{equation}\small
    \begin{aligned}\label{eq:2}
       \hat{I}_{q}(i)=\sum\nolimits_j A(i,j)~I_{r}(j).
    \end{aligned}
    \vspace{-1pt}
\end{equation}
The training objective of $\phi$ is hence defined as a reconstruction loss:
\vspace{-2pt}
\begin{equation}\small
    \begin{aligned}\label{eq:3}
       \mathcal{L}_{\text{res}}\!=\!||{I}_{q}-\hat{I}_{q}||_2.
    \end{aligned}
\vspace{1pt}
\end{equation}
In practice, to avoid trivial solutions caused by information leakage, an \textit{information bottleneck} is adopted over training samples,$_{\!}$ \eg,$_{\!}$ RGB2gray$_{\!}$ operation$_{\!}$~\cite{vondrick2018tracking},$_{\!}$ channel-wise$_{\!}$ drop- out in RGB$_{\!}$~\cite{lai2019self} or Lab$_{\!}$~\cite{lai2020mast} colorspace. After training,~the representation encoder $\phi$ is used for correspondence matching: similar to Eq.$_{\!}$~\ref{eq:2}, the affinity $A$ is estimated and used to \textit{propagate} desired pixel-level entities (\eg, instance masks, key-point maps), from a reference frame to a query frame.

\vspace{-1pt}
\subsection{\textbf{\textsc{Liir}}: \underline{L}ocality-Aware$_{\!}$ \underline{I}nter-and$_{\!}$ \underline{I}ntra-Video$_{\!}$ \underline{R}econstruction Framework}\label{sec:3.2}
\vspace{-1pt}
Building upon the reconstruction-by-copy scheme, \textsc{Liir} is empowered with three crucial yet long overlooked abilities for robust correspondence learning: instance discrimination, location awareness, and spatial compactness.

\begin{figure}[t]
	\vspace{-8pt}
	\begin{center}
		\includegraphics[width=0.99\linewidth]{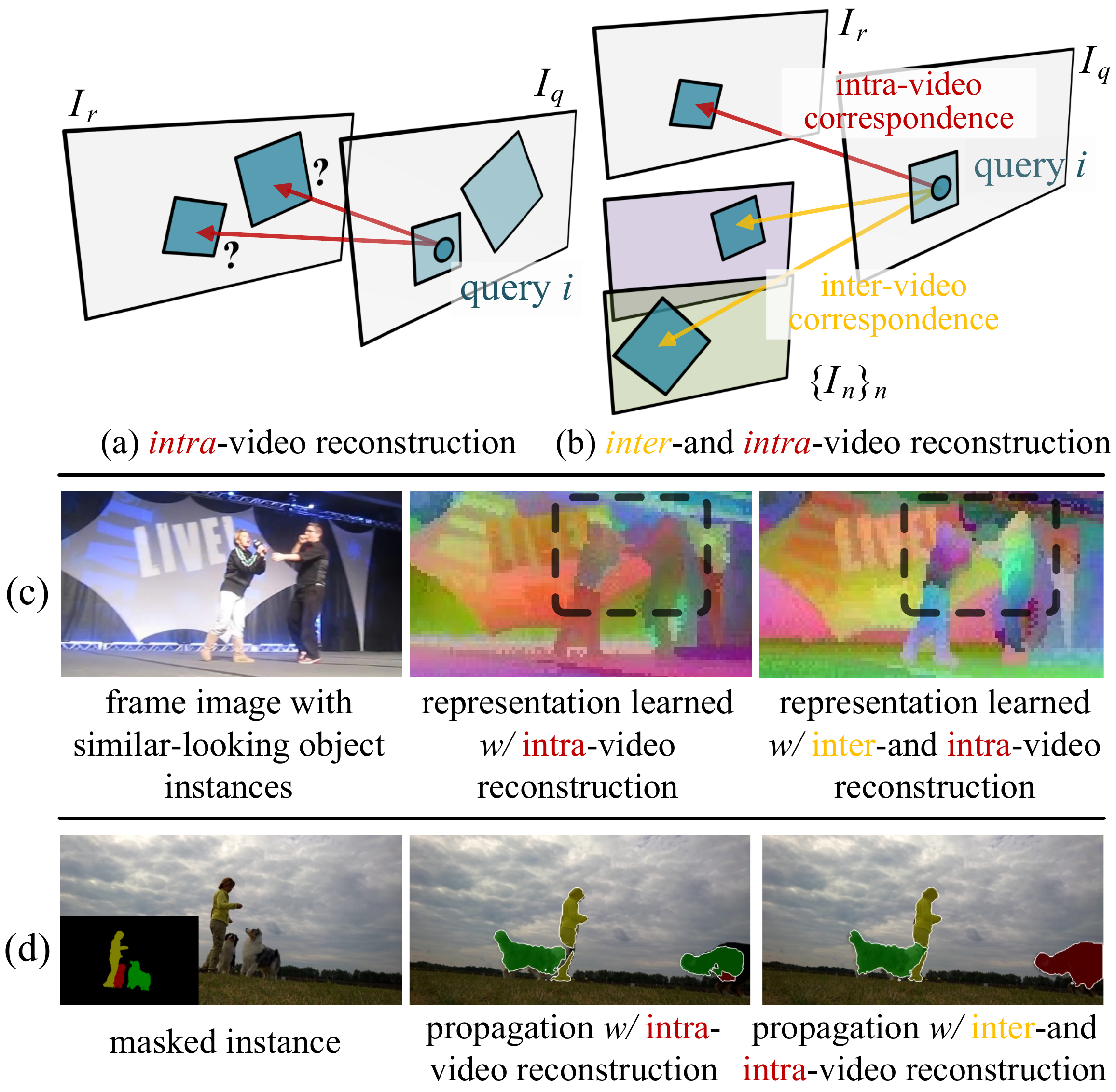}
\put(-80,208){\cmark}
\put(-73,188){\xmark}
\put(-83,158){\xmark}
		\end{center}
	\vspace{-17pt}
	\captionsetup{font=small}
	\caption{\small$_{\!}$\textbf{Inter-and$_{\!}$ intra-video$_{\!}$ reconstruction$_{\!}$}~(\S\ref{sec:3.2}).$_{\!}$ (a) Pre- vious$_{\!}$ intra-video$_{\!}$ reconstruction$_{\!}$ based$_{\!}$ approaches$_{\!}$ struggle$_{\!}$ to$_{\!}$ offer supervisory signal for distinguishing between different instances. (b) In our inter-and intra-video reconstruction, each query~pixel is forced to distinguish intra-video correspondence (\protect\includegraphics[scale=0.25,valign=c]{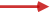}) from negative inter-video association (\protect\includegraphics[scale=0.25,valign=c]{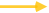}), enabling cross-instance discrimi- nation. (c)-(d) Representation learned with inter-and intra-video reconstruction is more robust for multiple object instances.
	}
	\label{fig:ivr}
	\vspace{-12pt}
\end{figure}

\noindent\textbf{Inter-and$_{\!}$ Intra-Video$_{\!}$ Reconstruction.$_{\!}$} With$_{\!}$ the$_{\!}$~computa- tion of the intra-video affinity $A$ (Eq.$_{\!}$~\ref{eq:1}), each query~pixel is forced to distinguish its counterpart (positive) reference pixels from unrelated (negative) ones \textit{within a same video}, with$_{\!}$ the$_{\!}$ indictor$_{\!}$ of$_{\!}$ the$_{\!}$ reconstruction$_{\!}$ quality$_{\!}$ $\mathcal{L}_{\text{res}\!}$  (Eqs.$_{\!}$~\ref{eq:2}-\ref{eq:3}).$_{\!\!}$\\
\noindent As both the positive and negative samples are sourced~from the same video, there is less evidence for distinction~among similar$_{\!}$  object$_{\!}$  instances$_{\!}$  with$_{\!}$ intra-video$_{\!}$ appearance$_{\!}$ only$_{\!}$ (Fig.$_{\!}$~\ref{fig:ivr}(a)).$_{\!}$ As$_{\!}$ one$_{\!}$ single$_{\!}$ video$_{\!}$ only$_{\!}$ contains$_{\!}$ limited content,$_{\!}$ conducting$_{\!}$ correspondence$_{\!}$ matching$_{\!}$ within$_{\!}$~videos$_{\!}$ is less challenging, and inevitably hinders the discrimination potential of the learned representation$_{\!}$~\cite{wang2021contrastive}. These insights motivate us to improve the intra-video affinity based reconstruction scheme by further accounting for negative \textit{across-video} correspondence. Concretely, given the query ($I_{q}$) and reference ($I_{r}$) frames  from the same video, an \textit{intra-inter video affinity} $A'_{\!}\!\in\![0,1]^{hw_{\!}\times_{\!}hw\!}$ is computed:
\vspace{-0pt}
\begin{equation}\small
    \begin{aligned}\label{eq:4}
\!\!\!\!\!\!A'(i,j)\!=\!\frac{\exp(\bm{I}_{q}(i)\!\cdot\!\bm{I}_{r}(j))}{\begingroup
      \color{ggray}
\underbrace{\color{black}\textstyle\sum\nolimits_{j'\!}\exp(\bm{I}_{q}(i)\!\cdot\!\bm{I}_{r}(j'))}_{\substack{\text{ \footnotesize {\color{myred}\textit{intra}}-video}\\ \text{ \footnotesize  correspondence}}} \endgroup
\!+_{\!}\!
\begingroup
      \color{ggray}
\underbrace{\color{black}\textstyle\sum\nolimits_{n}\!\sum\nolimits_{k\!}\exp(\bm{I}_{q}(i)\!\cdot\!\bm{I}_{n\!}(k))}_{\substack{\text{ \footnotesize {\color{myyellow}\textit{inter}}-video} \\ \text{\footnotesize correspondence}}}
\endgroup
},\!\!\!
    \end{aligned}
    \vspace{-2pt}
\end{equation}
where $\{{I}_{n}\}_n$ refer to a collection of frames, which are sampled from the whole training dataset, except the source video of $I_{q}$ ($I_{r}$). By additionally considering other irrelevant videos during affinity computation, both the quantity and diversity of negative samples are greatly improved, allowing us to derive a more challenging \textit{inter-and intra-video reconstruction} scheme (Fig.$_{\!}$~\ref{fig:ivr}(b)):
\vspace{-2pt}
\begin{equation}\small
    \begin{aligned}\label{eq:5}
       \hat{I}_{q}(i)=\sum\nolimits_j A'(i,j)~I_{r}(j).
    \end{aligned}
\vspace{-2pt}
\end{equation}
With Eqs.$_{\!}$~\ref{eq:4}-\ref{eq:5}, each pixel $i$ in the query frame ${I}_{q\!}$ is required to distinguish its counterpart pixels from massive unrelated ones, which are from not only the reference frame~${I}_{r}$ in current video, but a huge amount of irrelevant frames~$\{{I}_{n}\}_n$ in other videos. This powerful idea, yet, is elegantly achieved by the same training objective as in Eq.$_{\!}$~\ref{eq:3}. Note that Eq.$_{\!}$~\ref{eq:4} normalizes intra-video correspondence over both inter-and intra-video pixel-to-pixel relevance, while Eq.$_{\!}$~\ref{eq:5} only uses the pixels in the reference frame ${I}_{r}$ for reconstruction.  The rationale here is that, even if the encoder $\phi$ wrongly matches$_{\!}$ a$_{\!}$ query$_{\!}$ pixel$_{\!}$ $i$ with$_{\!}$ a$_{\!}$ negative$_{\!}$ but$_{\!}$ similar-looking$_{\!}$ pixel$_{\!}$ $k$ in ${I}_{n}$, \ie, $\exp(\bm{I}_{q}(i)_{\!}\cdot_{\!}\bm{I}_{n\!}(k))$ will be large and $\sum_jA'(i,j)\!\ll\!1$, the synthesized color $\hat{I}_{q}(i)$ will be \textit{still} very different to ${I}_{q}(i)$ and $\phi$ will receive a large gradient from Eq.$_{\!}$~\ref{eq:3}. Thus $\phi$ is driven to mine more high-level semantics and context-related clues, hence reinforcing the instance-level discrimination ability (Fig.$_{\!}$~\ref{fig:ivr}(c)). Fig.$_{\!}$~\ref{fig:ivr}(d) shows that the representation learned with our inter-and intra-video reconstruction strategy can distinguish similar-looking dogs nearby.

Although \cite{wang2021contrastive} also addresses inter-video analysis based reconstruction, it conducts embedding association on \textit{patch level}, relying on a pre-trained tracker for patch alignment. Besides, the method consumes three loss terms for supervision, which is much more complicated than ours. Further, it separately conducts inter-and intra-video affinity based reconstruction. This is problematic; when both the reference frame in current video and an irrelevant frame from other videos contain query-like pixels, there is no explicit supervision signal to determine which one should be matched.

\noindent\textbf{Position Encoding and Position Shifting.} Plenty of literature in neuroscience has revealed that human visual system encodes both appearance and position information when~we perceive$_{\!}$ and$_{\!}$ track$_{\!}$ objects$_{\!}$~\cite{livingstone1987psychophysical,hess1992coding,oksama2016position}.$_{\!}$ Yet$_{\!}$ existing$_{\!}$ unsuper- vised correspondence methods put all focus on improving appearance based representation by ConvNets, ignoring~the value$_{\!}$ of$_{\!}$ position$_{\!}$ information.$_{\!}$ Although$_{\!}$~\cite{islam2020much,kayhan2020translation}$_{\!}$~suggest that\\
\noindent  ConvNets$_{\!}$ can$_{\!}$ implicitly$_{\!}$ capture$_{\!}$ position$_{\!}$ information$_{\!}$ by$_{\!}$~uti- lizing$_{\!}$ image$_{\!}$ boundary$_{\!}$ effects,$_{\!}$ explicit$_{\!}$ position$_{\!}$ encoding$_{\!}$~has\\
\noindent already been a core of full$_{\!}$ attention$_{\!}$ networks$_{\!}$ (\eg, Trans- former$_{\!}$~\cite{DBLP:conf/nips/VaswaniSPUJGKP17}), and$_{\!}$ facilitated$_{\!}$ a$_{\!}$ variety$_{\!}$ of$_{\!}$ tasks$_{\!}$ (\eg, instance segmentation$_{\!}$ \cite{wang2020solov2}, tracking$_{\!}$ \cite{liu2020object}, video segmentation$_{\!}$ \cite{ci2018video}). All these indicate that position encoding deserves more attention in the field of temporal correspondence
learning.

Along this direction, \textsc{Liir} explicitly injects a position encoding map $\bm{P}\!\in\!\mathbb{R}^{h'{\!}\times_{\!}w'{\!}\times_{\!}c'\!}$ into the feature encoder $\phi$:
\vspace{-4pt}
\begin{equation}\small
    \begin{aligned}\label{eq:6}
\bm{I}=\phi(I, \bm{P}),
    \end{aligned}
\vspace{-3pt}
\end{equation}
where $\bm{P}$ is added with the output feature of the first conv layer of $\phi$ and  has the same size and dimension as
the conv feature. We explore three \textbf{position encoding strategies}:
\begin{itemize}[leftmargin=*]
	\setlength{\itemsep}{0pt}
	\setlength{\parsep}{-2pt}
	\setlength{\parskip}{-0pt}
	\setlength{\leftmargin}{-15pt}
	\vspace{-5pt}
\item \textit{2D Sinusoidal Position Embedding} (2DSPE): 
$\bm{P}$ is given with a family of pre-defined sinusoidal functions,  without introducing new trainable parameters:
\vspace{-6pt}
\begin{equation}\small
    \begin{aligned}\nonumber
\bm{P}(x,y,2u)&\!=\!\sin(x\!\cdot\! \varepsilon^{_{\!}\frac{4u}{c'}\!}),
\bm{P}(x,y,2u\!+_{\!}\!1_{\!})\!=\!\cos(x\!\cdot\! \varepsilon^{_{\!}\frac{4u}{c'}\!}),\\[-0.5ex]
\!\!\!\bm{P}(x,y,2v\!+_{\!}\!\frac{c'}{2}_{\!})&\!=\!\sin(y\!\cdot\! \varepsilon^{_{\!}\frac{4v}{c'}\!}),
\bm{P}(x,y,2v\!+_{\!}\!1_{\!}\!+_{\!}\!\frac{c'}{2}_{\!})\!=\!\cos(y\!\cdot\! \varepsilon^{_{\!}\frac{4v}{c'}\!}),\!
    \end{aligned}
\vspace{-8pt}
\end{equation}
where$_{\!}$ $x_{\!}\!\in_{\!}\![0,w')$,$_{\!}$ $y_{\!}\!\in_{\!}\![0,h')$$_{\!}$ specify$_{\!}$ the$_{\!}$ horizontal$_{\!}$ and vertical positions, $u,v_{\!}\!\in_{\!}\![0,c'/4)$ specify the dimension, and $\varepsilon_{\!}\!=_{\!}\!10^{-4}$. The horizontal (vertical)  positions are encoded in the first (second) half of the dimensions. 2DSPE naturally handles resolutions that are unseen  during training.
\item \textit{1D Absolute Position Embedding} (1DAPE): 1DAPE is the most heavy-weight strategy: the whole $\bm{P}$ is a learnable parameter matrix without any constraint.
\item \textit{2D Absolute Position Embedding} (2DAPE): As in$_{\!}$~\cite{dosovitskiy2020image}, two$_{\!}$ separate$_{\!}$ parameter$_{\!}$ sets:$_{\!}$ $\bm{X}_{\!}\!\in_{\!}\!\mathbb{R}^{w'_{\!\!\!}\!\times_{\!}c'_{\!\!\!\!}\!/{2}\!}$ and $\bm{Y}_{\!}\!\in_{\!}\!\mathbb{R}^{h'_{\!\!\!}\!\times_{\!}c'_{\!\!\!\!}\!/{2}}$, are learned for encoding the horizontal and vertical  positions, respectively, and combined to generate $\bm{P}$.
\vspace*{-3pt}
\end{itemize}

\begin{figure}[t]
	\vspace{-12pt}
	\begin{center}
		\includegraphics[width=0.90\linewidth]{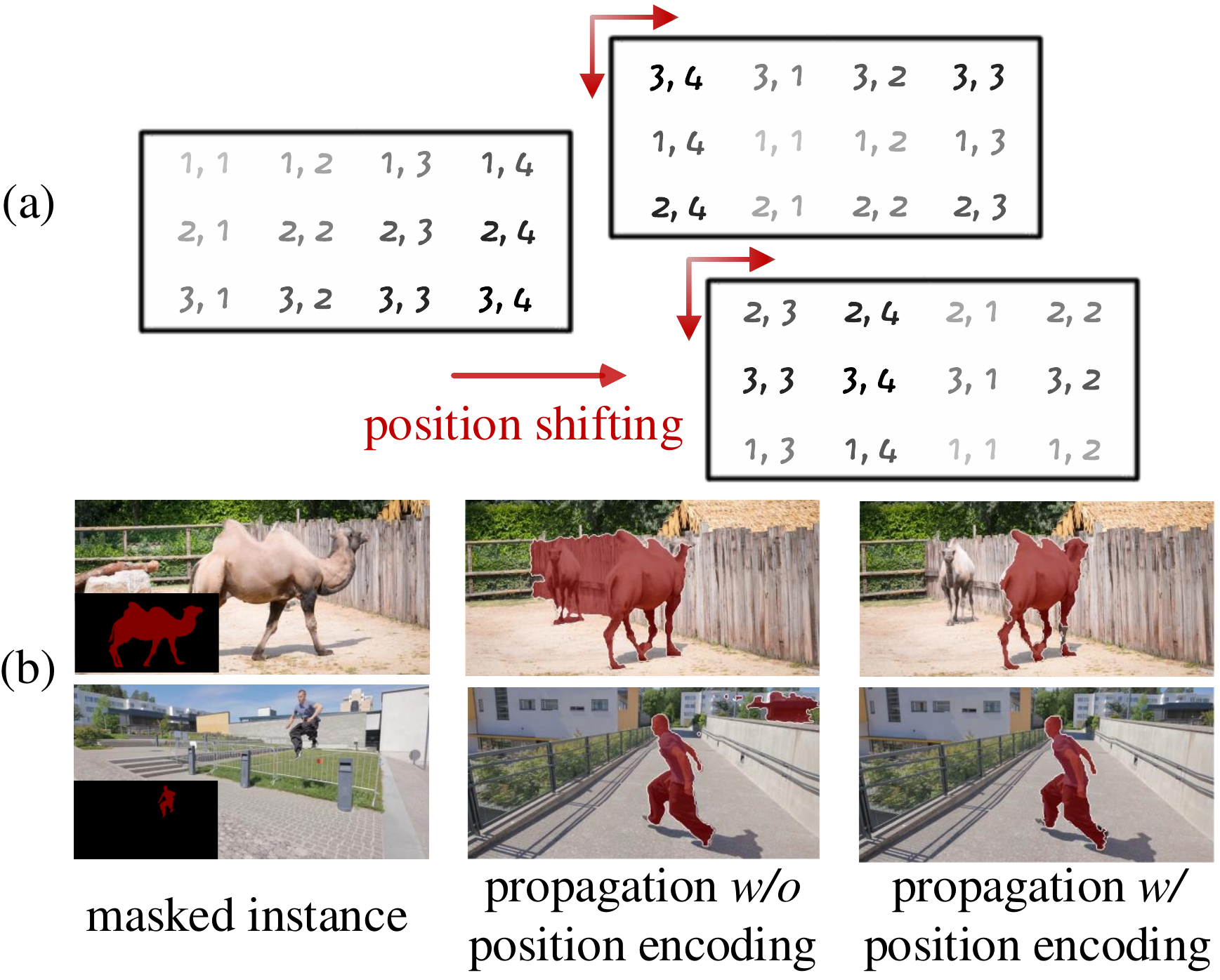}
        \put(-170,103){\small ${\bm{P}}$}
        \put(-30,128){\small $\bar{\bm{P}}$}
		\end{center}
	\vspace{-17pt}
	\captionsetup{font=small}
	\caption{\small (a) Illustration of \textbf{position shifting}. (b) Effect of \textbf{position encoding}. See \S\ref{sec:3.2} for details.
	}
	\label{fig:3}
	\vspace{-12pt}
\end{figure}

\noindent With$_{\!}$ our$_{\!}$ intra-inter$_{\!}$ video$_{\!}$ affinity$_{\!}$ (Eq.$_{\!}$~\ref{eq:4}),$_{\!}$  exploiting$_{\!}$ position\\
\noindent information$_{\!}$ in$_{\!}$ intra-video$_{\!}$ correspondence$_{\!}$ matching,$_{\!}$~\ie, $\{\exp(\bm{I}_{q}(i)\!\cdot\!\bm{I}_{r}(j))\}_j$, addresses local continuity resides in videos. However, for inter-video pixel relevance computation, \ie, $\{\exp(\bm{I}_{q}(i)\!\cdot\!\bm{I}_{n\!}(k))\}_{n,k}$, such position prior is undesirable, as it inspires the query pixel $i$ in $I_{q\!}$ to prefer matching these pixels with similar positions in other irrelevant videos $\{{I}_{n}\}_n$. To eliminate such  position encoding induced bias from inter-video correspondence matching, we design a \textbf{position shifting} strategy (Fig.$_{\!}$~\ref{fig:3}(a)). During training, for ${I}_{n\!}$ from other videos, we circularly shift the position encoding vectors in $\bm{P}$ by a random step in horizontal and vertical axes, respectively. The reason why we adopt random shifting with circular boundary conditions, instead of random shuffling, is to preserve the spatial layout in the modulated position encoding map $\bar{\bm{P}}$.  Then $\bar{\bm{P}}$ and ${I}_{n\!}$ are fed into $\phi$ for inter-video correspondence matching, and $\bar{\bm{P}}$ related gradients are abandoned if learnable 1DAPE or 2DAPE is used. Note$_{\!}$ that$_{\!}$ the$_{\!}$ standard$_{\!}$ position$_{\!}$ encoding$_{\!}$ $\bm{P}$$_{\!}$ is$_{\!}$ applied$_{\!}$ for$_{\!}$ the query (${I}_{q}$) and reference (${I}_{r}$) frames and updated normally. Fig.$_{\!}$~\ref{fig:3}(b) intuitively shows that merging position information into visual representation can enable robust correspondence matching even with confusing background and fast motion. In \S\ref{sec:abs}, we will quantitatively verify that 1DAPE is more favored and indeed boosts the performance.

\begin{figure}[t]
	\vspace{-8pt}
	\begin{center}
		\includegraphics[width=0.99\linewidth]{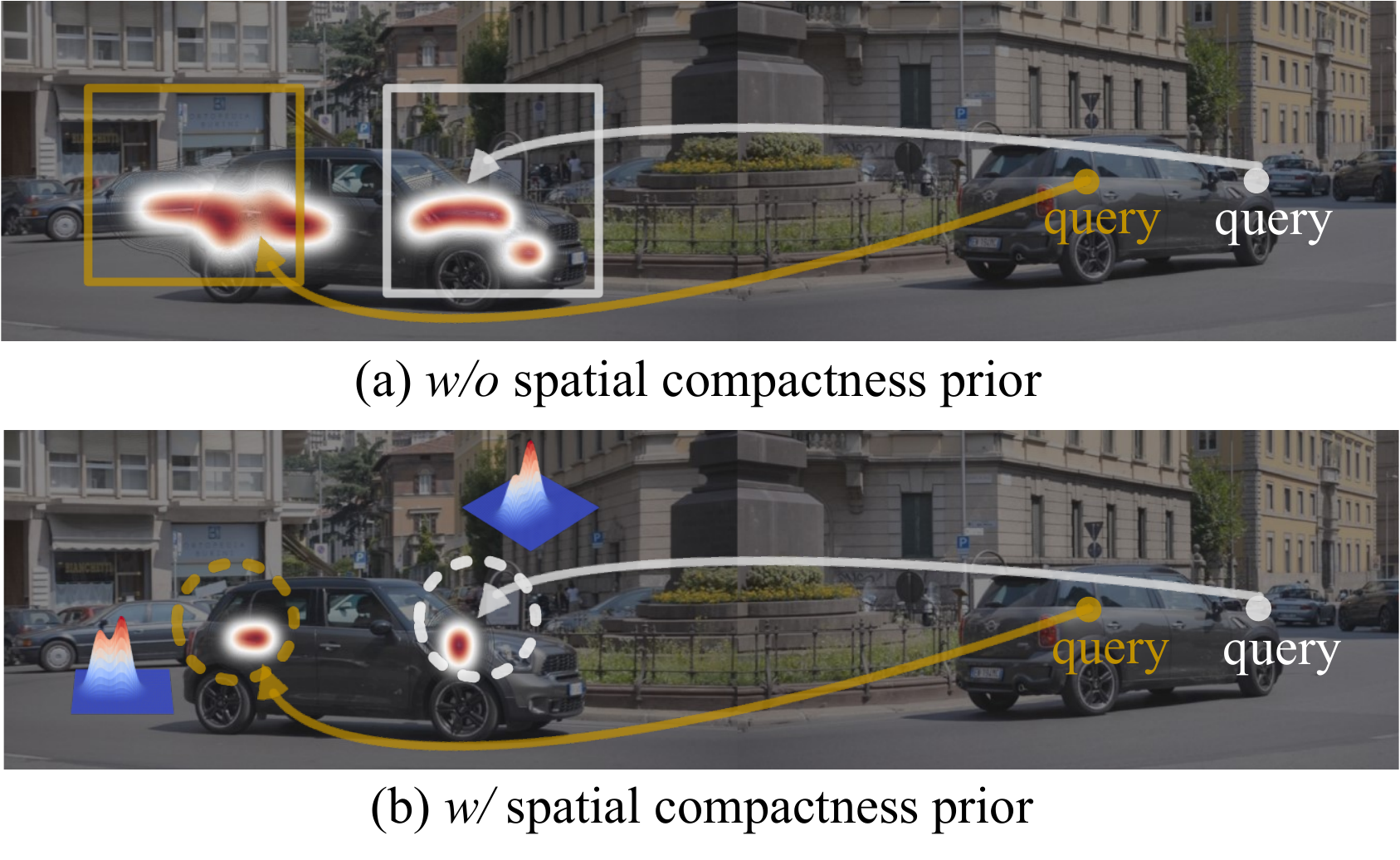}
		\end{center}
	\vspace{-17pt}
	\captionsetup{font=small}
	\caption{\small Illustration of \textbf{spatial compactness prior} (\S\ref{sec:3.2}).
	}
	\label{fig:scp}
	\vspace{-12pt}
\end{figure}

\begin{table*}[!t]
	\centering
	\small
	\resizebox{0.99\textwidth}{!}{
		\setlength\tabcolsep{1pt}
		\renewcommand\arraystretch{1.05}
		\begin{tabular}{cccrlccccc}
			\hline\thickhline
		    \rowcolor{mygray}
			Method & Backbone & Supervised & Dataset &(size) & $\mathcal{J}$ \& $\mathcal{F}$(Mean) $\uparrow$ & $\mathcal{J}$(Mean) $\uparrow$ & $\mathcal{J}$(Recall) $\uparrow$  &  $\mathcal{F}$(Mean) $\uparrow$  & $\mathcal{F}$(Recall) $\uparrow$  \\ \hline \hline
			Colorization\!~\cite{vondrick2018tracking} & ResNet-18 &  & Kinetics &(~-~, 800 hours)
			& 34.0 & 34.6 & 34.1 & 32.7 & 26.8 \\
			CorrFlow\!~\cite{lai2019self} & ResNet-18 &  & OxUvA &(~-~, 14 hours)
			& 50.3 & 48.4 & 53.2 & 52.2 & 56.0 \\
			TimeCycle\!~\cite{wang2019learning} & ResNet-50 &  & VLOG &(~-~, 344 hours)
			& 48.7 & 46.4 & 50.0 & 50.0 & 48.0 \\
			UVC\!~\cite{li2019joint} & ResNet-50 &  & C + Kinetics &(30k, 800 hours)
			& 60.9 & 59.3 & 68.8 & 62.7 & 70.9 \\
			MuG\!~\cite{lu2020learning} & ResNet-18 &   & OxUvA &(~-~, 14 hours)
			& 54.3 & 52.6 & 57.4 & 56.1 & 58.1  \\
			MAST\!~\cite{lai2020mast} & ResNet-18 &   & Youtube-VOS &(~-~, 5.58 hours)
			& 65.5 & 63.3 & 73.2 & 67.6 & 77.7  \\
			CRW\!~\cite{jabri2020space} & ResNet-18 &  & Kinetics &(~-~, 800 hours)
			& 68.3 & 65.5 & 78.6 & 71.0 & 82.9 \\
			ContrastCorr\!~\cite{wang2021contrastive} & ResNet-18 &  & C + TrackingNet &(30k, 300 hours)
			& 63.0 & 60.5 & - & 65.5 & - \\
			VFS \!~\cite{xu2021rethinking} & ResNet-18 &  & Kinetics &(~-~, 800 hours)
			& 66.7 & 64.0 & - & 69.4 & - \\
			JSTG\!~\cite{zhao2021modelling} & ResNet-18 &  & Kinetics &(~-~, 800 hours)
			& 68.7 & 65.8 & 77.7  & 71.6 & 84.3 \\
			CLTC\!~\cite{jeon2021mining}$^\dagger$  & ResNet-18 &  & Youtube-VOS &(~-~, 5.58 hours)
			& 70.3 & 67.9 & 78.2 & 72.6 & 83.7 \\
			DINO\!~\cite{caron2021emerging} & ViT-B/8 &  & I &(1.28M, ~-~) & 71.4 & 67.9 & - & \textbf{74.9} & - \\
			\textbf{\textsc{Liir}} & ResNet-18 &  & Youtube-VOS &(~-~, 5.58 hours)
			& \textbf{72.1} & \textbf{69.7} & \textbf{81.4} & 74.5 & \textbf{85.9} \\
			
			\hline
			\color{mygray2}ResNet\!~\cite{he2016deep} & \color{mygray2}ResNet-18 & \color{mygray2}\cmark & \color{mygray2}I &\color{mygray2}(1.28M, ~-~)
			& \color{mygray2}62.9 & \color{mygray2}60.6 & \color{mygray2}69.9 & \color{mygray2}65.2 & \color{mygray2}73.8 \\
			\color{mygray2}OSVOS\!~\cite{caelles2017one} & \color{mygray2}VGG-16 & \color{mygray2}\cmark & \color{mygray2}I+D &\color{mygray2}(1.28M, 10k)
			& \color{mygray2}60.3 & \color{mygray2}56.6 & \color{mygray2}63.8 & \color{mygray2}63.9 & \color{mygray2}73.8 \\
			\color{mygray2}FEELVOS\!~\cite{voigtlaender2019feelvos} & \color{mygray2}Xception-65 & \color{mygray2}\cmark & \color{mygray2}I + C + D + Youtube-VOS &\color{mygray2}(1.28M, 663k)
			& \color{mygray2}71.5 & \color{mygray2}69.1 & \color{mygray2}79.1 & \color{mygray2}74.0 & \color{mygray2}83.8 \\
			\color{mygray2}STM\!~\cite{oh2019video} & \color{mygray2}ResNet-50 & \color{mygray2}\cmark & \color{mygray2}I + D + Youtube-VOS &\color{mygray2}(1.28M, 164k)
			& \color{mygray2}81.8 & \color{mygray2}79.2 & \color{mygray2}88.7 & \color{mygray2}84.3 & \color{mygray2}91.8 \\ \hline
		\end{tabular}
	}
\leftline{~~\footnotesize{$^\dagger$: using task-specific model weights and architectures. ~~I: ImageNet$_{\!}$~\cite{deng2009imagenet}. ~~C: COCO$_{\!}$~\cite{lin2014microsoft}. ~~D: DAVIS$_{17}$\!~\cite{perazzi2016benchmark}.}}
	\captionsetup{font=small}
	\caption{\textbf{Quantitative results for video object segmentation} (\S\ref{sec:vos}) on DAVIS$_{17}$\!~\cite{perazzi2016benchmark} \texttt{val}.  For size of datasets, we report (\#\textit{raw} images, length of \textit{raw} videos) for self-supervised methods and (\#image-level annotations, \#pixel-level annotations) for {\color{mygray2}supervised} methods. }
	\label{table:davis}
	\vspace{-8pt}
\end{table*}

\begin{figure*}[t]
	\begin{center}
		\includegraphics[width=\linewidth]{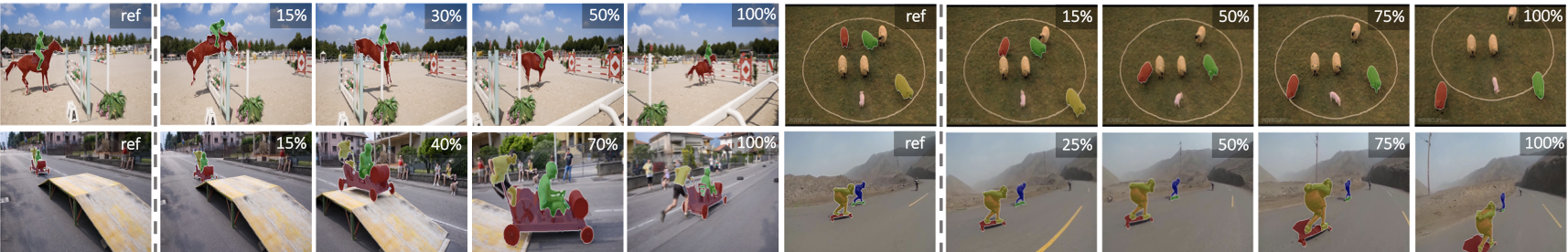}
	\end{center}
	\vspace{-18pt}
	\captionsetup{font=small}
	\caption{\small\textbf{Qualitative results for video object segmentation} (\S\ref{sec:vos}), on DAVIS$_{17}$$_{\!}$~\cite{perazzi2016benchmark} \texttt{val} (left) and Youtube-VOS\!~\cite{xu2018youtube} \texttt{val} (right).}
	\label{fig:results_davis}
	\vspace{-15pt}
\end{figure*}

\noindent\textbf{Spatial Compactness Prior.} As the visual world is conti- nuous and smoothly-varying, it is reasonable to assume appearances in video data change smoothly both in spatial~and\\
\noindent temporal$_{\!}$ dimensions.$_{\!}$ For$_{\!}$ correspondence$_{\!}$ learning,$_{\!}$ the$_{\!}$~tem- poral coherence has been extensively studied, while the spatial continuity received far less attention. To reduce search region, some existing methods$_{\!}$~\cite{vondrick2018tracking,lai2019self,lai2020mast} restrict corres- pondence matching within a local window, considering spatial regularities in a simple way. To make a better use of the spatial continuity, we augment the original reconstruction objective with an additional prior, termed as \textit{spatial compa- ctness}. Such a prior poses constraints on the spatial distribution of associated pixels, leading to \textit{sparse} and \textit{coherent} solutions. Specifically, given the query (${I}_{q}$) and reference (${I}_{r}$) frames, we expect \textbf{i)} each query pixel $i$ to be only matched with a small number of reference pixels, and \textbf{ii)} the matched reference pixels to be clustered. For a query pixel $i$ and its matching `heatmap':$_{\!}$ $A_{i\!\!}=_{\!\!}[A(i,j)]_{j\!}\!\in_{\!}\![0,1]^{h_{\!}\times_{\!}w}$, \wrt ${I}_{r}$, we assume $A_{i}$ follows a mixture of $M$ 2D Gaussian distributions:
\vspace{-4pt}
\begin{equation}\small
    \begin{aligned}\label{eq:7}
\mathcal{P}(x,y)=\sum\nolimits_{m=1}^M \omega_m~\mathcal{N}(x,y|\bm{\mu}_m,\bm{\Sigma}_m),
    \end{aligned}
\vspace{-3pt}
\end{equation}
where $(x, y)$ specifies the coordinate of a pixel location. We set $\{\bm{\mu}_{m\!}\!=_{\!}\![\mu_{x,m}, \mu_{y,m}]\}_{m\!}$ as the coordinates of top-$M$ scores in $A_{i}$, and set $M\!=\!2$ to address sparse robust matching. Other parameters, \ie, $\{\bm{\Sigma}_{m\!}\!=_{\!}\![\sigma^2_{x,m},0;0,\sigma^2_{y,m}]\}_m$, $\{\omega_m\}_m$, can be estimated efficiently from
$A_{i}$ without incurring high computational cost. In this way, we can derive a `compact' matching heatmap $\tilde{A}_{i\!}\!\in_{\!}\![0,1]^{h_{\!}\times_{\!}w\!}$ for each query pixel $i$, and eventually have $\tilde{A}_{\!}\!=_{\!}\![\tilde{A}_i]_{i\!}\!\in_{\!}\![0,1]^{hw_{\!}\times_{\!}hw}$. Such spatial compactness prior~$\tilde{A}$ is fully aware of \textbf{i)} and \textbf{ii)}, and used to regularize our representation learning:
\vspace{-3pt}
\begin{equation}\small
    \begin{aligned}\label{eq:8}
       \mathcal{L}_{\text{com}}\!=\!||\tilde{A}-{A}||_2.
    \end{aligned}
    \vspace{-2pt}
\end{equation}
Note that $\mathcal{L}_{\text{com}}$ is only applied for intra-video correspondence matching. Moreover, we replace ${A}$ with $\tilde{A}$ during inference, which eliminates outliers effectively. Two recent fully supervised video segmentation methods also explore the local continuity via~single-Gaussian locality prior~\cite{seong2020kernelized} or top-$k$ matching filtering~\cite{cheng2021modular}, while both of them can be viewed as specific instances of  our mixture-Gaussian based compactness prior, despite our different task settings. Fig.$_{\!}$~\ref{fig:scp} shows that our spatial compactness prior helps build reliable correspondence by inspiring sparse and compact solutions. Related experiments can be found in \S\ref{sec:abs}.

\subsection{Implementation Details}
\noindent\textbf{Network$_{\!}$ Configuration.}$_{\!}$ For$_{\!}$ fair$_{\!}$ comparison,$_{\!}$ our$_{\!}$ feature$_{\!}$ encoder$_{\!}$ $\phi$$_{\!}$ is$_{\!}$ implemented$_{\!}$ as$_{\!}$ ResNet-18\!~\cite{he2016deep} as in~\cite{vondrick2018tracking,jabri2020space,xu2021rethinking}. Following~\cite{lai2019self,lai2020mast,jeon2021mining}, $\times$2 downsampling is only made in the third residual block. Thus $\phi$ finally outputs 256 feature maps of
1/4 size of the input, \ie, $h_{\!}\!=_{\!}\!\frac{H}{4}, w_{\!}\!=_{\!}\!\frac{W}{4}, c_{\!}\!=_{\!}\!256$.  The position embedding is added to feature after the first 7$\times$7 \texttt{Conv}-\texttt{BN}-\texttt{ReLU} layer, \ie, $h'_{\!}\!=_{\!}\!\frac{H}{2}, w'_{\!}\!=_{\!}\!\frac{H}{2}, c'_{\!}\!=_{\!}\!64$.

\noindent\textbf{Training:} \textsc{Liir} is \textit{trained from scratch} {on two NVIDIA RTX-$3090$ GPUs} and only uses the raw
videos from Youtube-VOS\!~\cite{xu2018youtube}. Each training image is resized into 256$\times$256 and channel-wise dropout in Lab colorspace$_{\!}$~\cite{lai2020mast} is adopted as the information bottleneck.  Adam optimizer is used. At the initial $30$ epochs, only intra-video reconstruction learning is adopted for warm-up, with a learning rate of $10^{-3\!}$ and batch size of $32$. Then we conduct inter-video reconstruction learning with spatial compactness based regularization at the next $5$ epochs, with a learning rate of $10^{-4\!}$ and batch size of $12$. We online maintain a memory bank of 1,440 frames from different videos. For each query pixel, we sample 4 feature points from each stored frame, \ie, a total of 1,440$\times$4 negative samples are used for the inter-video correspondence computation, and we employ the moving average strategy~\cite{tarvainen2017mean,wu2018unsupervised,he2020momentum} for parameter updating.

\noindent\textbf{Testing:} Once \textsc{Liir} finishes training, there is no fine-tuning when applied to downstream tasks. Note that we utilize the compactness prior enhanced inter-frame affinity $\tilde{A}$ for mask propagation. As in \cite{oh2019video,lai2020mast}, we take multiple frames as reference for the full use of temporal context: at time step $t$, previous frames $I_0$, $I_5$, $I_{t-5}$, $I_{t-3}$, and $I_{t-1}$ (if applicable) are referred for current-frame mask propagation.

\begin{figure*}[t]
	\begin{center}
		\includegraphics[width=\linewidth]{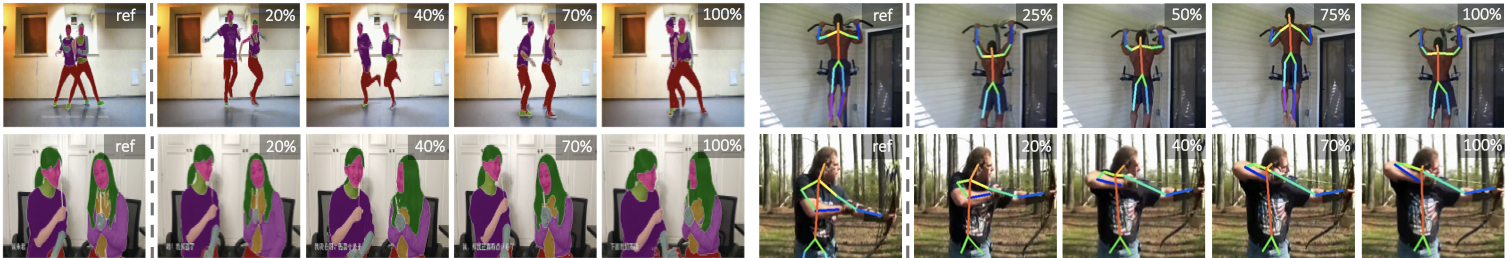}
	\end{center}
	\vspace{-18pt}
	\captionsetup{font=small}
	\caption{\small\textbf{$_{\!}$Qualitative$_{\!}$ results$_{\!}$ for$_{\!}$ part$_{\!}$ propagation}$_{\!}$ (\S\ref{sec:bps})$_{\!}$ and$_{\!}$ \textbf{pose$_{\!}$ tracking}$_{\!}$ (\S\ref{sec:hkp}),$_{\!}$ on$_{\!}$ VIP\!~\cite{zhou2018adaptive}$_{\!}$ \texttt{val}$_{\!}$ (left)$_{\!}$ and$_{\!}$ JHMDB\!~\cite{jhuang2013towards}$_{\!}$ \texttt{val}$_{\!}$ (right).$_{\!}$}
	\label{fig:results_vip}
	\vspace{-15pt}
\end{figure*}

\section{Experiment}
We evaluate the learned representation on diverse video label$_{\!}$ propagation$_{\!}$ tasks,$_{\!}$ \ie,$_{\!\!}$ video$_{\!}$ object$_{\!}$ segmentation$_{\!}$ (\S\ref{sec:vos}),$_{\!\!}$ body part propagation  (\S\ref{sec:bps}), and pose keypoint tracking (\S\ref{sec:hkp}). As in conventions~\cite{vondrick2018tracking,jabri2020space,xu2021rethinking}, all these tasks are to propagate the first frame annotation to the whole video sequence, and we use our model to
compute inter-frame dense correspondences. In \S\ref{sec:abs}, we conduct a set of ablative studies to examine the efficacy of our essential model designs.

\subsection{Results for Video Object Segmentation} \label{sec:vos}
\noindent\textbf{Dataset.} We first test our method on \texttt{val} sets of two popular video object segmentation datasets, \ie, DAVIS$_{17}$ \cite{perazzi2016benchmark} and YouTube-VOS \cite{xu2018youtube}.  There are $30$ and $474$ videos in DAVIS$_{17}$ and YouTube-VOS \texttt{val} sets, respectively.

\noindent\textbf{Evaluation Metric.} Following the official protocol~\cite{perazzi2016benchmark}, we use the region similarity ($\mathcal{J}$) and contour accuracy ($\mathcal{F}$) as the evaluation metrics. Note that the scores on YouTube-VOS are respectively reported for \textit{seen} and \textit{unseen} categories, obtained from the official evaluation server.

\begin{table}
	\centering
	\small
	\resizebox{\columnwidth}{!}{
		\setlength\tabcolsep{6pt}
		\renewcommand\arraystretch{1.05}
		\begin{tabular}{ccccccc}
			\hline\thickhline
		     \rowcolor{mygray}
		     & & & \multicolumn{2}{c}{Seen} & \multicolumn{2}{c}{Unseen} \\ \cline{4-7}
            \rowcolor{mygray}
            \multirow{-2}{*}{Methods} & \multirow{-2}{*}{Sup.} & \multirow{-2}{*}{Overall} & $\mathcal{J}$ $\uparrow$ & $\mathcal{F}$ $\uparrow$ & $\mathcal{J}$ $\uparrow$ & $\mathcal{F}$ $\uparrow$\\ \hline\hline
			Colorization\!~\cite{vondrick2018tracking} &  & 38.9 & 43.1 & 38.6 & 36.6 & 37.4 \\
			CorrFlow\!~\cite{lai2019self} &  & 46.6 & 50.6& 46.6 & 43.8 & 45.6 \\
			MAST\!~\cite{lai2020mast} &  & 64.2 & 63.9 & 64.9 & 60.3& 67.7 \\
			CLTC\!~\cite{jeon2021mining}$^\dagger$ &  & 67.3 & 66.2 & 67.9 & 63.2& 71.7 \\
			\textbf{\textsc{Liir}} &   & \textbf{69.3} & \textbf{67.9} & \textbf{69.7} & \textbf{65.7} & \textbf{73.8} \\
			\hline
			\color{mygray2}OSVOS\!~\cite{caelles2017one} & \color{mygray2}\cmark & \color{mygray2}58.8 & \color{mygray2}59.8& \color{mygray2}60.5 & \color{mygray2}54.2 & \color{mygray2}60.7 \\
			\color{mygray2}PreMVOS\!~\cite{luiten2018premvos} & \color{mygray2}\cmark & \color{mygray2}66.9 & \color{mygray2}71.4& \color{mygray2}75.9 & \color{mygray2}56.5 & \color{mygray2}63.7 \\
			\color{mygray2}STM\!~\cite{oh2019video} & \color{mygray2}\cmark & \color{mygray2}79.4 & \color{mygray2}79.7 & \color{mygray2}84.2 & \color{mygray2}72.8&\color{mygray2} 80.9
			\\ \hline
		\end{tabular}
	}
\leftline{~~\footnotesize{$^\dagger$: using task-specific model weights and architectures.}}
	\captionsetup{font=small}
	\caption{\small\textbf{Quantitative$_{\!}$ results$_{\!}$ for$_{\!}$ video$_{\!}$ object$_{\!}$ segmentation}$_{\!}$ (\S\ref{sec:vos}) on Youtube-VOS~\cite{xu2018youtube} \texttt{val}.}
	\label{table:youtube}
	\vspace{-15pt}
\end{table}

\noindent\textbf{Performance on DAVIS$_{17}$:} As illustrated in Table$_{\!}$~\ref{table:davis}, our \textsc{Liir} consistently outperforms all existing self-supervised methods across all the evaluation metrics. For example, it surpasses current best-performing self-supervised method, \ie, CLTC$_{\!}$~\cite{jeon2021mining}, in terms of mean $\mathcal{J}\&\mathcal{F}$ (\textbf{72.1} $_{\!}$\textit{vs.}$_{\!}$ 70.3). In addition, even without using \textit{any} manual annotations for training, \textsc{Liir} achieves very competitive segmentation performance in comparison with some famous supervised models~\cite{caelles2017one,voigtlaender2019feelvos} trained with massive pixel-wise annotations.

\noindent\textbf{Performance on YouTube-VOS \texttt{val}.}
Table~\ref{table:youtube} reports performance comparison of \textsc{Liir} against four self-supervised competitors on YouTube-VOS \texttt{val}. It can be observed that \textsc{Liir} sets new state-of-the-art. In particular, \textsc{Liir} yields an overall score of $\textbf{69.3\%}$, surpassing the second-best (\ie, CLTC$_{\!}$~\cite{jeon2021mining}) and third-best (\ie, MAST$_{\!}$~\cite{lai2020mast}) approaches by $\textbf{2.0\%}$ and $\textbf{5.1\%}$, respectively. Further, \textsc{Liir} even outperforms some famous supervised methods (\ie, OSVOS$_{\!}$~\cite{caelles2017one} and PreMVOS$_{\!}$~\cite{luiten2018premvos}), especially for the \textit{unseen} categories, clearly demonstrating its remarkable generalization ability.

\noindent\textbf{Qualitative Results.}
Fig.$_{\!}$~\ref{fig:results_davis} depicts visual results on representative videos in the datasets. As seen, \textsc{Liir} is able to establish accurate correspondences under various challenging scenarios, \eg, scale changes, small objects and occlusions.

\subsection{Results for Body Part Propagation} \label{sec:bps}

\noindent\textbf{Dataset.} We next evaluate our model performance for body part propagation. Experiments are conducted on VIP \texttt{val}$_{\!}$~\cite{zhou2018adaptive}, which contains 50 videos with annotations of $19$ human semantic part categories (\eg, hair, face, dress).

\noindent\textbf{Evaluation Metric.} As suggested by VIP~\cite{zhou2018adaptive}, we adopt {mean intersection-over-union} (mIoU) and {mean Average Precision} (mAP) metrics for evaluation of semantic-level and instance-level parsing, respectively.

\noindent\textbf{Performance.} As shown in Table~\ref{table:vip}, \textsc{Liir} achieves state-of-the-art performance  on both semantic-level and instance-level parsing. This indicates that \textsc{Liir} can generate strong representation which models both cross-instance discrimination and intra-instance invariance well.  Fig.$_{\!}$~\ref{fig:results_vip} depicts some visualization results on two representative videos. \textsc{Liir} achieves temporally stable results and shows robustness to typical challenges (\eg, pose variations, occlusions).

\begin{table}
	\centering
	\small
	\resizebox{\columnwidth}{!}{
			\setlength\tabcolsep{3pt}
		\renewcommand\arraystretch{1.05}
		\begin{tabular}{cccc||cc}
            \hline\thickhline
		    \rowcolor{mygray}
		     & & \multicolumn{2}{c||}{VIP} & \multicolumn{2}{c}{JHMDB} \\ \cline{3-6}
            \rowcolor{mygray}
            \multirow{-2}{*}{Methods} & \multirow{-2}{*}{Sup.} & mIoU $\uparrow$ & AP $\uparrow$ & PCK@0.1 $\uparrow$ & PCK@0.2 $\uparrow$
			\\ \hline \hline
			TimeCycle\!~\cite{wang2019learning} &  & 28.9 & 15.6 & 57.3 & 78.1 \\
			UVC\!~\cite{li2019joint} &  & 34.1 & 17.7 & 58.6 & 79.6\\
			CRW\!~\cite{jabri2020space} &  & 38.6 & - & 59.3 & 80.3 \\
			ContrastCorr\!~\cite{wang2021contrastive} &  & 37.4 & 21.6 & 61.1 & 80.8 \\
			VFS \!~\cite{xu2021rethinking} &  & 39.9 & - & 60.5 & 79.5\\
			CLTC\!~\cite{jeon2021mining}$^\dagger$ &  & 37.8 & 19.1 & 60.5 & 82.3\\
			JSTG\!~\cite{zhao2021modelling} &  & 40.2 & - & \textbf{61.4} & \textbf{85.3}\\
			\textbf{\textsc{Liir}} &  & \textbf{41.2} & \textbf{22.1} & 60.7 & 81.5\\
		    \hline
			\color{mygray2}ResNet\!~\cite{he2016deep} & \color{mygray2}\cmark & \color{mygray2}31.9 & \color{mygray2}12.6 & \color{mygray2}53.8 & \color{mygray2}74.6\\
			\color{mygray2}ATEN\!~\cite{song2017thin} & \color{mygray2}\cmark & \color{mygray2}37.9 & \color{mygray2}24.1 & \color{mygray2}- & \color{mygray2}-\\
			\color{mygray2}TSN\!~\cite{zhou2018adaptive} & \color{mygray2}\cmark & \color{mygray2}- & - & \color{mygray2}68.7 & \color{mygray2}92.1\\
			\hline
		\end{tabular}
	}
\leftline{~~\footnotesize{$^\dagger$: using task-specific model weights and architectures.}}
	\captionsetup{font=small}
	\caption{\small\textbf{Quantitative results for part propagation} (\S\ref{sec:bps}) and \textbf{pose tracking} (\S\ref{sec:hkp}), on VIP$_{\!}$~\cite{zhou2018adaptive} \texttt{val} and JHMDB$_{\!}$~\cite{jhuang2013towards} \texttt{val}.}
	\label{table:vip}
	\vspace{-15pt}
\end{table}

\begin{table*}[t]
	\vspace{-1.5em}
    \hspace{0.1em}
    \renewcommand\thetable{5}
    \subfloat[{Inter-and Intra-Video Recons.} \label{table:ablation2}]{
        \resizebox{0.235\textwidth}{!}{
        \tablestyle{1pt}{1.1}
        		\begin{tabular}{c|c|c||c}
        \hline\thickhline
		     \rowcolor{mygray}
		    &\#Negative & \multicolumn{1}{c||}{DAVIS} & \multicolumn{1}{c}{VIP} \\ \cline{3-4}
            \rowcolor{mygray}
            \multirow{-2}{*}{\#}&\multirow{-1}{*}{Samples} & $\mathcal{J}$\&$\mathcal{F}_m$ $\uparrow$  & mIoU $\uparrow$
            \\ \hline\hline
			1&0   & 69.2 & 38.4 \\ \hline
            2&480 & 70.9 \color{mygreen}{(\textbf{+1.7})} & 39.8 \color{mygreen}{(\textbf{+1.4})} \\
            3&960 & 71.7 \color{mygreen}{(\textbf{+2.5})}& 41.0 \color{mygreen}{(\textbf{+2.6})}\\
			4&1,440  & \textbf{72.1} \color{mygreen}{(\textbf{+2.9})} & \textbf{41.2} \color{mygreen}{(\textbf{+2.8})}\\
			\hline
		\end{tabular}
        }
        }
    \subfloat[{Position Encoding}\label{table:ablation3}]{%
        \tablestyle{1pt}{1.1}
        \resizebox{0.215\textwidth}{!}{
        \begin{tabular}{c|c||c}
        \hline\thickhline
		     \rowcolor{mygray}
		    Position & \multicolumn{1}{c||}{DAVIS} & \multicolumn{1}{c}{VIP} \\ \cline{2-3}
            \rowcolor{mygray}
            Encoding & $\mathcal{J}$\&$\mathcal{F}_m$ $\uparrow$  & mIoU $\uparrow$
            \\ \hline\hline
			\textit{w/o} PE   &  70.9 &  40.3 \\ \hline
            2DSPE   & 70.6  \color{mywarning}{(\textbf{-0.3})}& 40.2  \color{mywarning}{(\textbf{-0.1})}\\
            1DAPE & \textbf{72.1} \color{mygreen}{(\textbf{+1.2})} & \textbf{41.2} \color{mygreen}{(\textbf{+0.9})}  \\
            2DAPE & 71.9  \color{mygreen}{(\textbf{+1.0})}& 41.1 \color{mygreen}{(\textbf{+0.8})}\\
			\hline
		\end{tabular}
        }}
    \subfloat[{Position Shifting}\label{table:ablation4}]{%
    \resizebox{0.231\textwidth}{!}{
        \tablestyle{1pt}{1.1}
        		\begin{tabular}{c|c||c}
        \hline\thickhline
		     \rowcolor{mygray}
		    Position & \multicolumn{1}{c||}{DAVIS} & \multicolumn{1}{c}{VIP} \\ \cline{2-3}
            \rowcolor{mygray}
            Modulation & $\mathcal{J}$\&$\mathcal{F}_m$ $\uparrow$  & mIoU $\uparrow$
            \\ \hline\hline
			N/A & 71.3 & 40.6 \\\hline
            shuffling & 71.8 \color{mygreen}{(\textbf{+0.5})} & 41.0 \color{mygreen}{(\textbf{+0.4})} \\
            shifting   & \textbf{72.1} \color{mygreen}{(\textbf{+0.8})} & \textbf{41.2} \color{mygreen}{(\textbf{+0.6})}\\
            \hline

            \multicolumn{3}{c}{}\\
		\end{tabular}
}}
    \subfloat[{Spatial Compactness Prior}\label{table:ablation5}]{%
    \resizebox{0.285\textwidth}{!}{
        \tablestyle{1pt}{1.1}
        		\begin{tabular}{cc|c||c}
        \hline\thickhline
		     \rowcolor{mygray}
		    \multicolumn{2}{c|}{Spatial Compactness} & \multicolumn{1}{c||}{DAVIS} & \multicolumn{1}{c}{VIP} \\ \cline{3-4}
            \rowcolor{mygray}
            \textit{training} &\textit{inference} & $\mathcal{J}$\&$\mathcal{F}_m$ $\uparrow$  & mIoU $\uparrow$
            \\ \hline\hline
             & & 69.8 & 39.6 \\\hline
            \cmark &  & 71.5 \color{mygreen}{(\textbf{+1.7})}& 40.8 \color{mygreen}{(\textbf{+1.2})}\\
            &\cmark   & 70.8\color{mygreen}{(\textbf{+1.0})} & 40.3 \color{mygreen}{(\textbf{+0.7})}\\
            \cmark &\cmark   & \textbf{72.1} \color{mygreen}{(\textbf{+2.3})} & \textbf{41.2} \color{mygreen}{(\textbf{+1.6})}  \\
			\hline
		\end{tabular}
}}
    \captionsetup{font=small}
    \caption{\small \textbf{A set of ablation studies} on DAVIS$_{17}$ \cite{perazzi2016benchmark} \texttt{val} and VIP$_{\!}$~\cite{zhou2018adaptive} \texttt{val}. See \S\ref{sec:abs} for details.}
    \label{tab:ablations}\vspace{-5mm}
\end{table*}

\subsection{Results for Pose Keypoint Tracking} \label{sec:hkp}

\noindent\textbf{Dataset.} We then examine the model performance on human keypoint tracking, using JHMDB\!~\cite{jhuang2013towards} \texttt{val}. JHMDB \texttt{val} has $268$ videos. For each person, a total of $15$ body joints, \eg, torso, head, shoulder, elbow, are annotated.

\noindent\textbf{Evaluation Metric.}
We use \textit{probability of correct keypoint} (PCK)~\cite{yang2012articulated} to measure the accuracy between each tracking result and corresponding ground-truth with a threshold $\tau$.

\noindent\textbf{Performance.} Table~\ref{table:vip} shows that \textsc{Liir} exhibits compelling overall performance. Note that CLTC\!~\cite{jeon2021mining} uses different checkpoints and model architectures for different tasks and datasets, while we only use a single model for evaluation. The visual results in Fig.$_{\!}$~\ref{fig:results_vip} also demonstrate the strong capability of \textsc{Liir} in establishing precise correspondence.

\subsection{Diagnostic Experiment} \label{sec:abs}
For further detailed analysis, we conduct a series of ablative studies on DAVIS$_{17}$ \cite{perazzi2016benchmark} \texttt{val} and VIP$_{\!}$~\cite{zhou2018adaptive} \texttt{val} sets.

\begin{table}
    \renewcommand\thetable{4}
	\centering
	\small
	\resizebox{\columnwidth}{!}{
		\tablestyle{3pt}{1.1}
		\begin{tabular}{c|ccc|c||c}
        \hline\thickhline
		     \rowcolor{mygray}
		     &Inter-and Intra-&Position & \multicolumn{1}{c|}{Spatial}& \multicolumn{1}{c||}{DAVIS} & \multicolumn{1}{c}{VIP} \\ \cline{5-6}
            \rowcolor{mygray}
            \multirow{-2}{*}{\#}&\multirow{1}{*}{Video Recons.} & \multirow{1}{*}{Encoding} & Compactness & $\mathcal{J}$\&$\mathcal{F}_m$ $\uparrow$  & mIoU $\uparrow$  \\ \hline\hline
			1& &  &    & 65.3 & 35.2\\ \hline
            2&\cmark &  & & 68.7 \color{mygreen}{(\textbf{+3.4})}& 38.4 \color{mygreen}{(\textbf{+3.2})}\\
            3& & \cmark &  & 66.9 \color{mygreen}{(\textbf{+1.6})}&  37.0 \color{mygreen}{(\textbf{+1.8})}\\
            4& &  & \cmark & 68.4 \color{mygreen}{(\textbf{+3.1})}&  37.2 \color{mygreen}{(\textbf{+2.0})} \\
			5&\cmark & \cmark & \cmark  & \textbf{72.1} \color{mygreen}{(\textbf{+6.8})} & \textbf{41.2} \color{mygreen}{(\textbf{+6.0})}\\
			\hline
		\end{tabular}
	}
	\captionsetup{font=small}
	\caption{\small \textbf{Detailed analysis} of essential components of \textsc{Liir} on DAVIS$_{17}$ \cite{perazzi2016benchmark} \texttt{val} and VIP$_{\!}$~\cite{zhou2018adaptive} \texttt{val}. See \S\ref{sec:abs} for details.}
	\label{table:ablation1}
	\vspace{-15pt}
\end{table}

\noindent\textbf{Key Component Analysis.} We first examine the efficacy of essential components of \textsc{Liir}, \ie, inter-and intra-video reconstruction, position encoding, and spatial compactness. The results are summarized in Table$_{\!}$~\ref{table:ablation1}, where position encoding is implemented as 1DAPE, and compactness prior is used during both training and inference stages. When separately comparing row \#2 - \#4 with the baseline (MAST$_{\!}$~\cite{lai2020mast})\\
\noindent in row \#1, we can observe that each individual module in-  deed$_{\!}$ boosts$_{\!}$ the$_{\!}$ performance.$_{\!}$ For$_{\!}$ example,$_{\!}$ on$_{\!}$ DAVIS$_{17}$$_{\!}$~\texttt{val},$_{\!}$ inter-and intra-video reconstruction, position encoding, and spatial  compactness prior respectively bring \textbf{3.4\%}, \textbf{1.6\%}, and \textbf{3.1\%} $\mathcal{J}\&\mathcal{F}$ gains. This verifies our core insight that these three elements are crucial for correspondence learning. Finally, in {row \#5}, we combine all the three components together -- \textsc{Liir}, and obtain the best performance. This suggests that these modules are complementary to each other, and confirms the effectiveness of our whole design.

\noindent\textbf{Inter-and Intra-Video Reconstruction.} We next study the impact of increasing the number  of negative samples, \ie, frames from other irrelevant videos used for inter-video correspondence computation (Eq.$_{\!}$~\ref{eq:4}). In Table$_{\!}$~\ref{table:ablation2}, row \#1 gives scores of learning without considering inter-video correspondence. In this case, the results are unsatisfactory. When more negative samples are involved (\ie, 0$\rightarrow$1,440), better performance can be achieved (\ie, 69.2$\rightarrow$72.1 on DAVIS$_{17}$ \texttt{val}, 38.4$\rightarrow$41.2 on VIP \texttt{val}).  Finally we use 1,440 negative samples for inter-video reconstruction based learning, which is the maximum number allowed by our GPU.

\noindent\textbf{Position Encoding.} To determine the effect of our position encoding module, we then report the performance with different encoding strategies in Table$_{\!}$~\ref{table:ablation3}. As seen, the non-learnable strategy, 2DSPE, hinders the performance, while the learnable alternatives, \ie, 1DAPE and 2DAPE, lead to better results. Compared with 2DAPE, 1DAPE is more favored, probably due to its high flexibility and capacity.

\noindent\textbf{Position Shifting.} We further study the influence of our position shifting strategy on performance.  As shown in Table$_{\!}$~\ref{table:ablation4}, we consider two alternatives, \ie, `NAN' and `position shuffling'. `NAN' refers to using the normal position encoding map $\bm{P}$ without any modulation during inter-video correspondence matching. Compared with `position shifting', `NAN' suffers from performance degradation (\ie, 72.1$\rightarrow$71.3 on DAVIS$_{17}$ \texttt{val}, 41.2$\rightarrow$40.6 on VIP \texttt{val}), showing the negative effect of the position-induced bias. The other baseline, `position shuffling', \ie, randomly shuffling the position encoding map for inter-video affinity computation, though better than `NAN', is still worse than  `position shifting'. This is because it destroys the spatial layouts.

\noindent\textbf{Spatial Compactness Prior.} The spatial compactness prior (Eq.$_{\!}$~\ref{eq:7}) is used to regularize intra-video correspondence matching during both training and inference stages. In Table$_{\!}$~\ref{table:ablation5}, we quantitatively identify the performance contribution of our spatial compactness prior in different stages.

\vspace{-5pt}
\section{Conclusion}
\vspace{-1pt}
We presented a self-supervised temporal correspondence learning approach, \textsc{Liir}, that makes contributions in three aspects. First, going beyond the popular intra-video analysis based learning scheme, we further enforce separation between intra- and inter-video pixel associations, enhancing instance-level feature discrimination. Second, with a clever position shifting strategy, we bring the advantages of position encoding into full play, while avoiding its undesirable impact at the same time. Third, a spatial compactness prior is introduced to regularize representation learning and improve correspondence inference. The effectiveness was thoroughly validated over various label propagation tasks.

{\small
\bibliographystyle{ieee_fullname}
\bibliography{egbib}
}

\clearpage

\appendix

\section{Analysis of Spatial Compactness Prior} \label{sc:scp}
We use $M$ 2D Gaussian distributions to approximate the affinity matrix $A$ between two frames. Table~\ref{table:ablation1} provides a detailed analysis of the hyper-parameter $M$. The $1$st row corresponds to a baseline model that disregards the spatial compactness prior in both training and inference phases. We see from the table that \textbf{1)} when $M\!=\!1$ in training, the model performs worse than the baseline model, as such a rigorous matching constraint easily leads to overconfident pre-dictions; \textbf{2)} when $M$ (in training) becomes larger, the performance greatly improves; \textbf{3)} the models trained with $M\!=\!2$ shows consistently better performance than those trained with $M\!=\!3$; and \textbf{4)} in the inference stage, $M\!=\!2$ always leads to the best performance. Accordingly, we set $M$ to $2$ in both training and inference stages.

\vspace{5pt}
\section{Analysis of Long-Term Dependence} \label{sc:scp2}
\textsc{Liir} leverages multiple reference frames during testing as in \cite{lai2020mast}. We analysis the impact of long-term dependence in Table \ref{table:long-term}. As seen, our method is more robust and shows smaller drop \textit{wrt} \cite{lai2020mast}: 4.7\% \textit{vs} 6.8\%.

\vspace{5pt}
\section{Analysis of Feature Point Sampling} \label{sc:scp3}
We give the ablative study on the number of feature points sampled during inter-video reconstruction in Table \ref{table:sampling}. It can be seen that, with 1440 frames, better results can be achieved if we sample more points per frame (70.4$\rightarrow$ 71.4$\rightarrow$72.1), supporting our claim about instance separation.  With the same number of total sampled points, we gain better performance if we consider more frames (70.9$\rightarrow$ 71.7$\rightarrow$72.1). It is reasonable as more frames can provide much rich/challenging context. With our limited GPU capacity, we choose to sample 1440 frames and four feature points per frame, so as to maximize the performance. But we can speculate that, if with enough GPU capacity, sampling more features points from more videos will further improve the performance.

\vspace{5pt}
\section{Visualization of Ablation Study }\label{sc:ablative}
\begin{figure*}[t]
    \renewcommand\thefigure{D.1}
	\vspace{-9pt}
	\begin{center}
		\includegraphics[width=\linewidth]{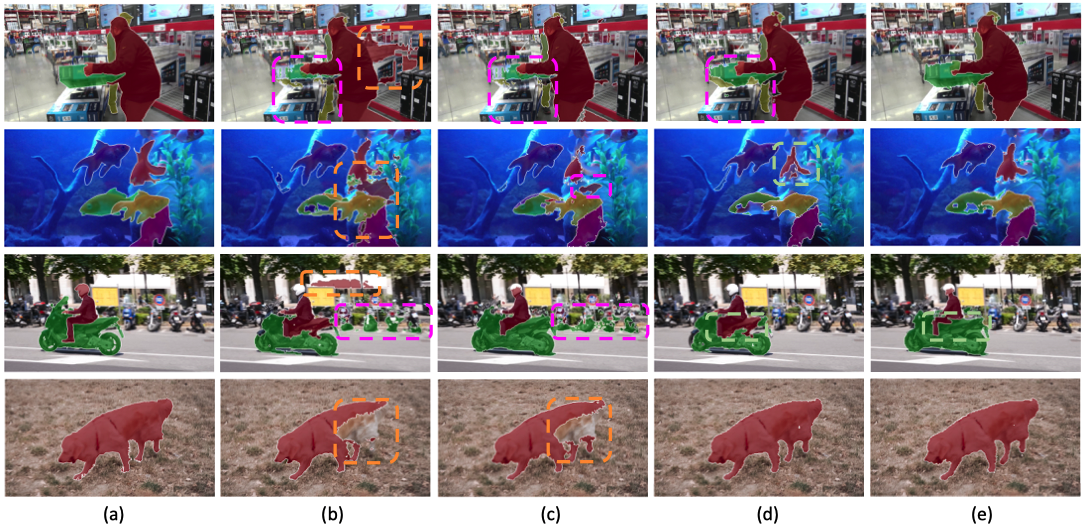}
		\end{center}
	\vspace{-17pt}
	\captionsetup{font=small}
	\caption{\small \textbf{Visual effects of essential components} in \textsc{Liir}. (a) ground truth, (b) baseline, (c) b + position encoding, (d) c + inter-video reconstruction, and (e) d + spatial compactness. See \S\ref{sc:ablative} for details.
	}
	\label{fig:ablation}
	\vspace{-14pt}
\end{figure*}

Fig.$_{\!}$~\ref{fig:ablation} depicts visual effect of each essential component in \textsc{Liir}. Starting from the baseline model (b), we progressively add position encoding (c), inter-video reconstruction (d) and spatial compactness (e). As seen, with explicit positional encoding, our model is able to heavily suppress background regions (\eg, shelves in the first row). The inter-video reconstruction enables more accurate discrimination between targets and semantically similar distractors  (\eg, ``motorcycle'' in the third row). Last, incorporating the spatial compactness prior facilitates more precise correspondences,  leading to high-quality segmentation results.
\begin{table}
    \renewcommand\thetable{A.1}
	\centering
	\small
	\resizebox{\columnwidth}{!}{
		\tablestyle{6pt}{1}
		\begin{tabular}{c|cc|c||c}
        \hline\thickhline
		     \rowcolor{mygray}
		     &&& \multicolumn{1}{c||}{DAVIS} & \multicolumn{1}{c}{VIP} \\ \cline{4-5}
            \rowcolor{mygray}
            \multirow{-2}{*}{\#}&\multirow{-2}{*}{training stage} & \multirow{-2}{*}{inference stage} &  $\mathcal{J}$\&$\mathcal{F}_m$ $\uparrow$  & mIoU $\uparrow$  \\ \hline\hline
			1& -- & -- &  69.8 & 39.6\\  \hline
			2& $M=1$ & -- &  69.4 & 39.3\\
			3& $M=1$ & $M=1$ &  69.0 & 39.1\\
			4& $M=1$  & $M=2$  &  69.6 & 39.4\\
			5& $M=1$  & $M=3$  &  69.6 & 39.4\\
			6& $M=2$  & --  &  71.5 & 40.8\\
			7& $M=2$  & $M=1$  &  71.2 & 40.4\\
			\rowcolor{myorange}
			8& $M=2$  & $M=2$  &  \textbf{72.1} & \textbf{41.2}\\
			9& $M=2$  & $M=3$  &  71.9 & \textbf{41.2}\\
			10& $M=3$ & --  &  71.3 & 40.7\\
            11& $M=3$ & $M=1$  &  71.1 & 40.4\\
            12& $M=3$ & $M=2$  &  71.7 & 40.9\\
            13& $M=3$ & $M=3$  &  71.6 & 40.8\\
			\hline
		\end{tabular}
	}
	\captionsetup{font=small}
	\caption{\small \textbf{Detailed analysis} of $M$ in spatial compactness prior on DAVIS$_{17}$ \cite{perazzi2016benchmark} \texttt{val} and VIP$_{\!}$~\cite{zhou2018adaptive} \texttt{val}. ``--'': without using spatial compactness prior. See \S\ref{sc:scp} for details.}
	\label{table:ablation1}
		\vspace{-6pt}
\end{table}

\vspace{-10pt}
\section{Additional Qualitative Results} \label{sc:vis}
\begin{figure*}[t]
    \renewcommand\thefigure{D.2}
	\vspace{-9pt}
	\begin{center}
		\includegraphics[width=\linewidth]{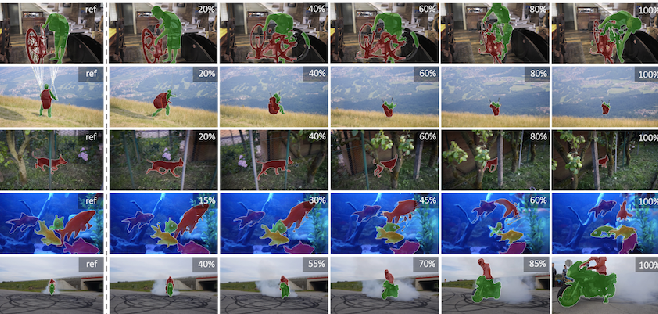}
		\end{center}
	\vspace{-17pt}
	\captionsetup{font=small}
	\caption{\small\textbf{More visualization results for video object segmentation} on DAVIS$_{17}$$_{\!}$~\cite{perazzi2016benchmark} \texttt{val}.}
	\label{fig:davis}
	\vspace{-14pt}
\end{figure*}

\begin{figure*}[t]
    \renewcommand\thefigure{D.3}
	\vspace{-9pt}
	\begin{center}
		\includegraphics[width=\linewidth]{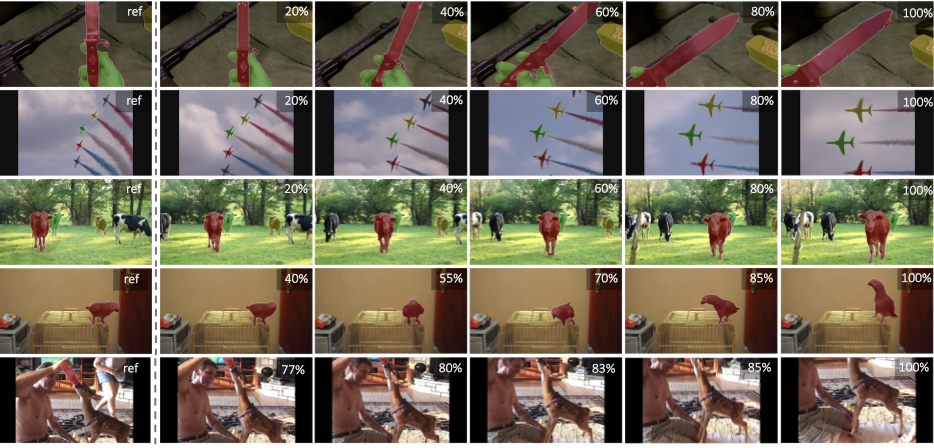}
		\end{center}
	\vspace{-17pt}
	\captionsetup{font=small}
	\caption{\small\textbf{More visualization results for video object segmentation} on Youtube-VOS\!~\cite{xu2018youtube} \texttt{val}.}
	\vspace{-14pt}
	\label{fig:ytb}
\end{figure*}

\begin{figure*}[t]
    \renewcommand\thefigure{D.4}
	\vspace{-9pt}
	\begin{center}
		\includegraphics[width=\linewidth]{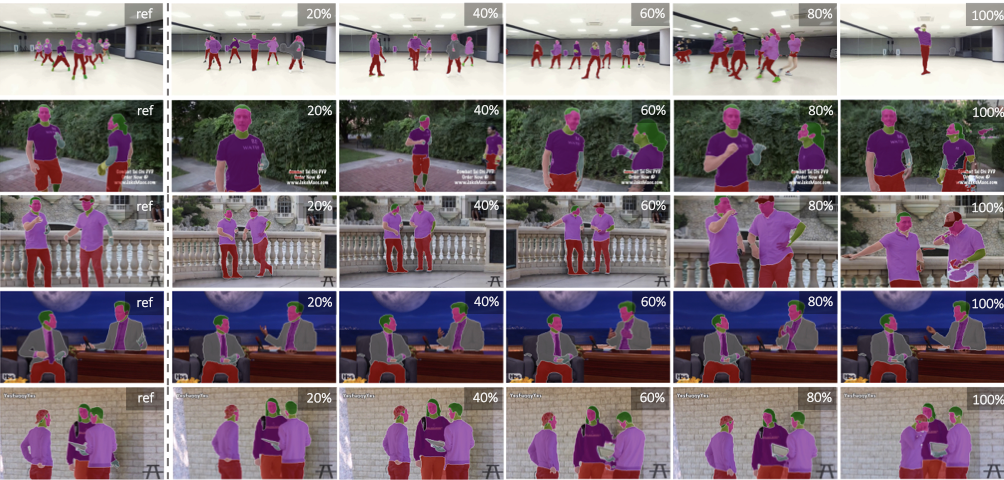}
		\end{center}
	\vspace{-17pt}
	\captionsetup{font=small}
	\caption{\small\textbf{More visualization results for body part propagation} on VIP$_{\!}$~\cite{zhou2018adaptive} \texttt{val}.}
	\label{fig:vip}
	\vspace{-14pt}
\end{figure*}

\begin{figure*}[t]
    \renewcommand\thefigure{D.5}
	\vspace{-9pt}
	\begin{center}
		\includegraphics[width=\linewidth]{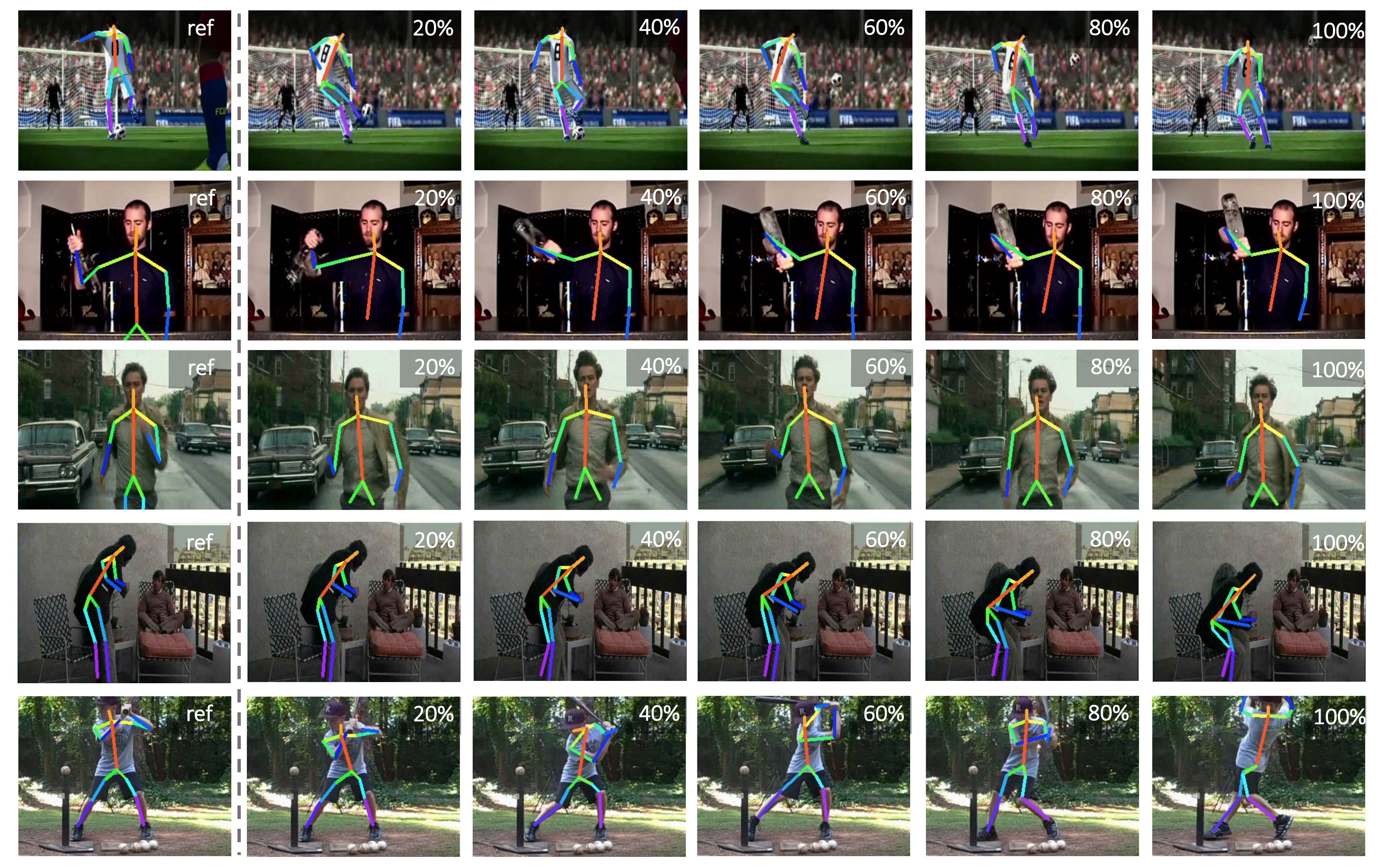}
		\end{center}
	\vspace{-17pt}
	\captionsetup{font=small}
	\caption{\small\textbf{More visualization results for keypoint propagation} on JHMDB$_{\!}$~\cite{jhuang2013towards} \texttt{val}.}
	\vspace{-14pt}
	\label{fig:jhmdb}
\end{figure*}

We provide additional video propagation results on four datasets, including DAVIS$_{17}$\!~\cite{perazzi2016benchmark} \texttt{val} in Fig.$_{\!}$~\ref{fig:davis}, Youtube-VOS\!~\cite{xu2018youtube} \texttt{val} in Fig.$_{\!}$~\ref{fig:ytb}, VIP\!~\cite{zhou2018adaptive} \texttt{val} in Fig.$_{\!}$~\ref{fig:vip} and JHMDB\!~\cite{jhuang2013towards} \texttt{val}  in Fig.$_{\!}$~\ref{fig:jhmdb}. We observe that even training with no annotations, \textsc{Liir} is able to produce highly exquisite results.

\begin{table}[b]
    \renewcommand\thetable{B.1}
	\centering
	\small
	\resizebox{0.99\columnwidth}{!}{
		\setlength\tabcolsep{6pt}
		\renewcommand\arraystretch{1.0}
		\begin{tabular}{c|c||c}
			\hline\thickhline
			\rowcolor{mygray}
			Methods & {Reference Frame} & $\mathcal{J}$\&$\mathcal{F}_m$ \\ \hline\hline
			Ours & {\color{mygray2}{\st{$I_0$}}}, {\color{mygray2}{\st{$I_5$}}}, $I_{t-5}$, $I_{t-3}$, $I_{t-1}$ & 72.1 $\rightarrow$ 67.4 {\color{mywarning}{(\textbf{-4.7})}} \\
			MAST\!~\cite{lai2020mast} & {\color{mygray2}{\st{$I_0$}}}, {\color{mygray2}{\st{$I_5$}}}, $I_{t-5}$, $I_{t-3}$, $I_{t-1}$ & 65.5 $\rightarrow$ 58.7 {\color{mywarning}{(\textbf{-6.8})}} \\
			\hline
		\end{tabular}
	}
	\captionsetup{font=small}
	\caption{\small Analysis of long-term dependence of reference frames on DAVIS$_{17}$ \cite{perazzi2016benchmark} \texttt{val}. See \S\ref{sc:scp2} for details..}
	\label{table:long-term}
	\vspace{-4pt}
\end{table}

\begin{table}[b]
    \renewcommand\thetable{C.1}
	\centering
	\small
	\resizebox{0.99\columnwidth}{!}{
		\setlength\tabcolsep{2pt}
		\renewcommand\arraystretch{1.0}
		\begin{tabular}{c||c|c|c|c|c}
			\hline\thickhline
			\cellcolor[gray]{0.9}\#Frames & 1440 & 1440 & 1440 & 960 & 480\\ \hline
			\cellcolor[gray]{0.9}\#Feature Points & \multirow{2}{*}4 & \multirow{2}{*}2 & \multirow{2}{*}1 & \multirow{2}{*}6 & \multirow{2}{*}{12}  \\
			\cellcolor[gray]{0.9}Per-frame & & & & &  \\\hline
            \cellcolor[gray]{0.9}$\mathcal{J}$\&$\mathcal{F}_m$ & \textbf{72.1}   &  71.4 {\color{mywarning}{(\textbf{-0.7})}} & 70.4 {\color{mywarning}{(\textbf{-1.7})}} & 71.7 {\color{mywarning}{(\textbf{-0.4})}} & 70.9 {\color{mywarning}{(\textbf{-1.2})}} \\
			\hline
		\end{tabular}
	}
		\captionsetup{font=small}
	\caption{\small Ablative study on the number of feature points sampled during inter-video reconstruction on DAVIS$_{17}$ \cite{perazzi2016benchmark} \texttt{val} (see \S\ref{sc:scp3}).}
	\label{table:sampling}
	\vspace{-3pt}
\end{table}

\section{Limitation} \label{sc:lim}
Although \textsc{Liir} demonstrates remarkable performance and high generalizability in correspondence matching, we still see a large performance gap between \textsc{Liir} and current top-leading supervised models (\eg, STM \cite{oh2019video} in VOS).  However, as a self-supervised method, \textsc{Liir} can be easily scaled to leverage any available collection of video data for training. This could lead to more accurate correspondence learning from massive unlabeled data instead of using small-scale datasets only (\eg, DAVIS, YouTube-VOS). Apart from that, inter-video reconstruction spares massive space to bank the negative samples, this coerces the memory size of GPUs and extends the training time. Fortunately, the performance of \textsc{Liir} has improved by leaps and bounds to compensate for it, for instance, up to \textbf{2.9\%} on DAVIS$_{17}$\!~\cite{perazzi2016benchmark}. Further more, note that the inter-video reconstruction is only applied during training, thus it does not introduce extra computational cost at inference.
In our future work, we will explore towards the above direction to narrow the performance gap between supervised methods, and find the way to reduce the memory space taken up by negative videos and speed up training simultaneously.

\newpage

\end{document}